\documentclass[10pt,a4paper]{article}
\usepackage[vmargin=2.2 cm, hmargin = 2.2 cm]{geometry}
\usepackage{amsmath}
\usepackage{amssymb}
\usepackage{amsthm}
\usepackage{bm}
\usepackage{epsfig}
\usepackage{epstopdf}
\usepackage{graphicx,xcolor}
\usepackage{soul} 
\usepackage{subfigure}
\usepackage{multirow}
\usepackage{url}
\usepackage{algorithm}
\usepackage{algorithmicx}
\usepackage{algpseudocode}  

\usepackage{comment}
\usepackage{booktabs}
\usepackage{enumitem}

\newcommand{\specialcell}[2][c]{\begin{tabular}[#1]{@{}c@{}}#2\end{tabular}}

\date{}

\begin{document}
\title{\texttt{LPC-AD}: Fast and Accurate  Multivariate Time Series Anomaly Detection via Latent Predictive Coding}

\author{
 Zhi Qi$^\ast$, Hong Xie$^\ast$, Ye Li$^\dag$, Jian Tan$^\dag$, FeiFei Li$^\dag$, John C.S. Lui$^\ddag$ \\
$^\ast$Chongqing University, $^\dag$Alibaba, $^\ddag$The Chinese University of Hong Kong\\
$^\ast$\{okjamesqi,xiehong2018\}\@cqu.edu.cn \\
$^\dag$\{liye.li, j.tan, lifeifei\}@alibaba-inc.com\\
$^\ddag$cslui@cse.cuhk.edu.hk 
} 

\maketitle

\begin{abstract}
This paper proposes \texttt{LPC-AD}, 
a fast and accurate multivariate time 
series (MTS) anomaly detection method.  
\texttt{LPC-AD} is motivated by the ever-increasing needs 
for fast and accurate MTS anomaly detection methods 
to support fast troubleshooting 
in cloud computing, micro-service systems, etc.  
\texttt{LPC-AD} is fast in the sense that its reduces the training time by as high 
as 38.2\% compared to the state-of-the-art (SOTA) deep learning methods 
that focus on training speed.  
\texttt{LPC-AD} is accurate in the sense that it improves  
the detection accuracy by as high as 18.9\%  
compared to SOTA sophisticated deep learning methods 
that focus on enhancing detection accuracy.  
Methodologically, \texttt{LPC-AD} 
contributes a generic architecture \texttt{LPC-Reconstruct} 
for one to attain different trade-offs 
between training speed and detection accuracy.  
More specifically, \texttt{LPC-Reconstruct} is built on ideas 
from autoencoder for reducing redundancy in time series, 
latent predictive coding for capturing temporal dependence in MTS,   
and randomized perturbation for avoiding overfitting 
of anomalous dependence in the training data.  
We present simple instantiations of \texttt{LPC-Reconstruct} 
to attain fast training speed, 
where we propose a simple randomized perturbation method.  
The superior performance of \texttt{LPC-AD} over SOTA methods is validated  
by extensive experiments on four large real-world datasets.  
Experiment results also show  
the necessity and benefit of each component of the 
\texttt{LPC-Reconstruct} architecture 
and that \texttt{LPC-AD} is 
robust to hyper parameters.
\end{abstract}

\section{{\bf Introduction}}

Multivariate time series serves as an important 
reference of the ``health status'' of many Internet applications 
such cloud computing, micro-service systems and the Internet routing network, to name a few.  
For example, consider the cloud computing system, 
which has become a crucial infrastructure for many customized 
services of companies and governments  \cite{ma2020diagnosing,meng2020localizing,liu2021microhecl}.  
To maintain a high quality of service of the cloud computing system, 
it is important to monitor the ``health status'' of the cloud system 
and perform troubleshooting in real-time \cite{liu2021microhecl,ma2020diagnosing,meng2020localizing}.  
To achieve this, many performance metrics of a cloud computing system 
such as CPU usage, disk I/O rate, network packet loss rate, 
etc., are monitored in real-time, 
which form a MTS \cite{liu2021microhecl,ma2020diagnosing,meng2020localizing}.   
From the MTS, system anomalies can be detected via machine learning    
\cite{audibert2020usad,Huang2020,Li2021,Malhotra2016,tuli2022tranad,zhang2019deep}.   
which greatly improves troubleshooting of the system.  

Real-world applications like cloud computing, 
micro-service systems, etc., generate large amount and high dimensional 
time series data and they needed to be processed by fast and accurate 
MTS anomaly detection methods.  
Although many machine learning algorithms were proposed to detect anomalies 
in MTS  \cite{audibert2020usad,Huang2020,Li2021,Malhotra2016,patcha2007overview,tuli2022tranad,zhang2019deep}, 
how to achieve a fast training speed while retaining fairly high detection 
accuracy is underexplored.  
Classical methods like  \cite{bandaragoda2014efficient,kingsbury2020elle,liu2008isolation,patcha2007overview,wang2020real,yaacob2010arima} 
have a fast training speed, but their detection accuracy is not high, 
due to low expressiveness capability of their model.  
Using deep learning models, methods like 
\cite{dai2021sdfvae,li2021multivariate,park2018multimodal,su2019robust} 
break the records of classical methods with 
a surprisingly high detection accuracy  
thanks to high expressiveness of these models.  
In the research line of modern methods, 
most efforts were spent on developing sophisticated 
models to improve detection accuracy 
\cite{dai2021sdfvae,li2021multivariate,park2018multimodal,su2019robust}.  
As a consequence, these sophisticated models becomes more and more  
time-consuming for training.  
Recently, the USAD \cite{audibert2020usad} and TranAD \cite{tuli2022tranad} 
improve the training speed, 
while retaining a fairly high accuracy.  
These two algorithms follow the same architecture of adversarial training 
(refer to Section \ref{sec:relatedwork} for details) and their detection accuracy 
can be low in some datasets (refer to Section \ref{sec:Experiment} for details).  
This paper aims to explore faster training speed 
and more accurate MTS anomaly detection methods.

It is challenging to design fast and accurate 
MTS anomaly detection algorithms.  
First, the MTS training data is usually not associated with anomalous labels 
and there is no specification on the pattern of an anomaly in general 
\cite{audibert2020usad,tuli2022tranad}.  
Second, the anomalous data points in MTS training data can mislead the trained model \cite{li2021multivariate}.  
Third, the MTS data is usually of high dimension and 
it has complex spatial and temporal dependence 
\cite{audibert2020usad,tuli2022tranad}, 
where the spatial dependence refers to the dependence among different time series.  
This makes it difficult to learn the normal and abnormal pattern 
from the MTS.  
Fourth, the MTS data is usually non-stationary, 
\cite{liu2021microhecl}
making it difficult to learn stable patterns.  
In this paper, we propose an algorithm called \texttt{LPC-AD} to 
address these challenges.  

The \texttt{LPC-AD} has a shorter training time 
than SOTA deep learning methods 
that focus on reducing training time, and it 
has a higher detection accuracy than 
SOTA sophisticated deep learning based methods that 
focus on enhancing detection accuracy.  
These merits of \texttt{LPC-AD} are supported by novelty in the design 
and extensive empirical evaluation on four large real-world datasets.  

\noindent
{\bf Novelty in the design.}  
Though the \texttt{LPC-AD} falls into the research line of 
autoencoder based methods \cite{audibert2020usad,Li2021,tuli2022tranad,zhang2019deep,Huang2020} 
that learn the normal spatial and temporal 
dependence in the MTS data to detect anomaly
(refer to Section \ref{sec:relatedwork} for details on this research line), 
it contributes several new ideas.  
Unlike previous works 
\cite{audibert2020usad,Li2021,tuli2022tranad,zhang2019deep,Huang2020} 
which use autoencoder to capture spatial and 
temporal dependence in many MTS, 
we use autoencoder to reduce redundancy in the time series data.  
Through this we obtain a low dimensional latent representation of the data.  
Evidence of  redundancy in MTS can be found in  \cite{liu2021microhecl,ma2020diagnosing,meng2020localizing}.  
As our purpose is to reduce redundancy, 
simple autoencoders such as vanilla LSTM based autoencoder suffice.  
To capture temporal dependence in MTS, 
we apply ideas from predictive coding \cite{atal1970adaptive}, 
which have been successfully applied to 
word representation\cite{mikolov2013efficient}, 
image colorization\cite{zhang2016colorful}, 
and visual representation learning \cite{doersch2015unsupervised}, etc.  
Note that we are the first to explore predictive coding 
for MTS anomaly detection.  
More specifically, we use a predictor to capture temporal dependence 
between two consecutive sliding windows of the MTS.  
The predictor aims to learn the normal dependence via  
predicting future latent variables from 
its preceding latent variables in the sliding window.  
Here, we do not need sophisticated prediction networks, 
as they may overfit the anomalous patterns in the training data.  
We then use randomized perturbation to inject some noise 
on the output of the predictor.  
Finally, we use the decoder to decode this perturbed prediction 
and use the absolute error between the decoded data and  
the ground truth time series data to measure 
the anomalous score of a data point.  
The purpose of this perturbation is to avoid overfitting 
to the patterns of anomaly data points, 
as the decoder aims to learn the normal patterns.
This randomized perturbation approach has a theoretical foundation 
in online learning in the presence of adversarial data \cite{Abernethy2016}.  
As an analogy, the anomalous data points in the time series 
are equivalent to the adversarial data in online learning.  
Note that this randomized perturbation only incurs a 
negligible computational cost.    
Combining them all, we contribute a generic architecture called 
\texttt{LPC-Reconstruct} in the sense that 
one can select different instances of 
the above mentioned autoencoder network, 
the predictor network and the randomized perturbation method 
to attain different tradeoffs between the training speed  
and detection accuracy.  

\begin{figure}[htb]
	\centering
	\includegraphics[width=0.8\textwidth]{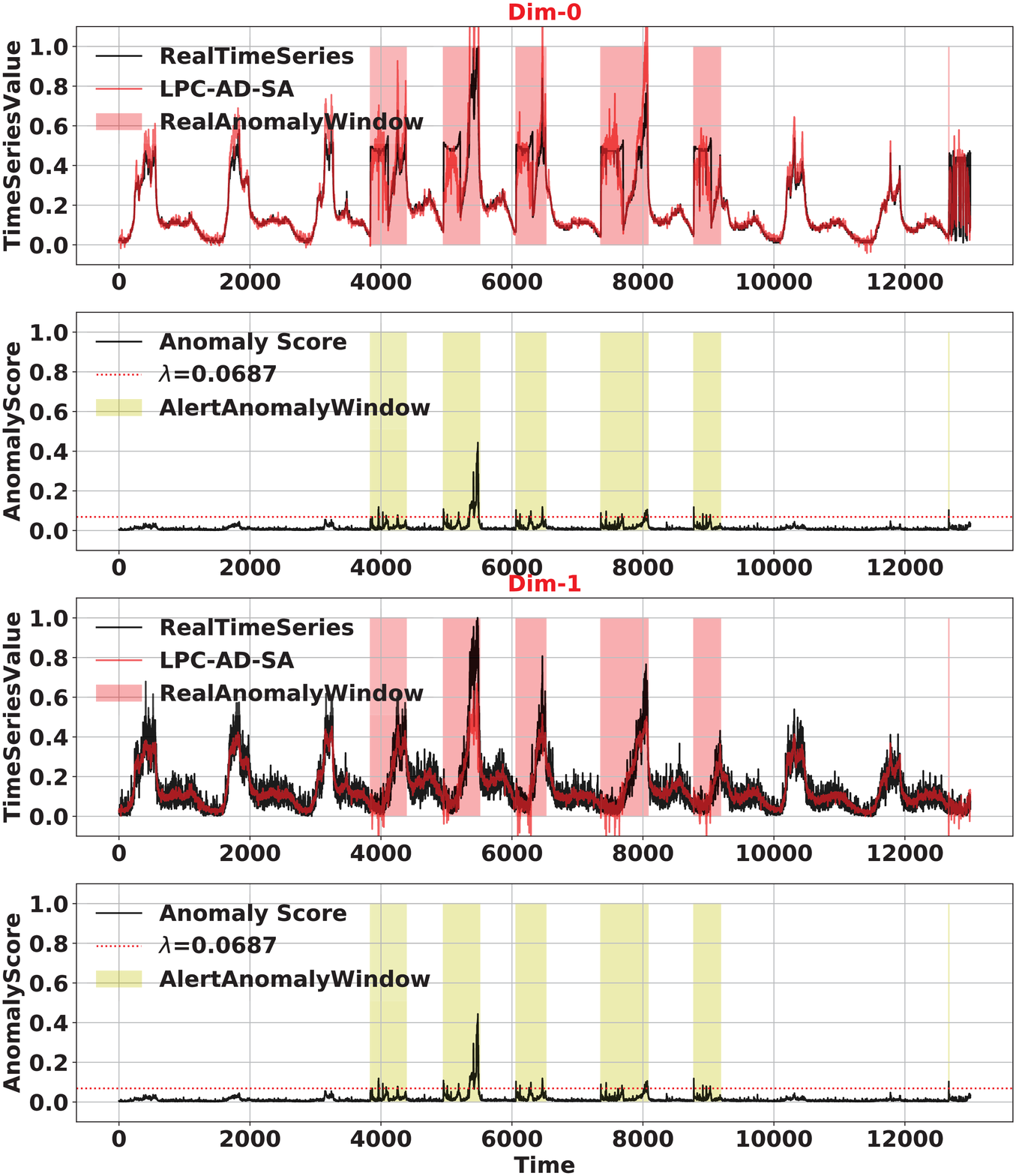}
	\caption{Visualizing the anomalies detected by \texttt{LPC-AD-SA}.}
	\label{fig:Imagescase}
	\vspace{-0.12in}
\end{figure}

To illustrate, Figure \ref{fig:Imagescase} visualizes anomalies detected 
by \texttt{LPC-AD-SA} in the SMD dataset, 
where the \texttt{LPC-AD-SA} denotes an instance of 
\texttt{LPC-AD} (refer to Section \ref{sec:Experiment} for more details).  
Only two dimensions of the time series are shown for simplicity.  
In Figure \ref{fig:Imagescase}, the red curve corresponds 
to reconstructions by \texttt{LPC-AD-SA}.  
The anomaly score is quantified by the absolute reconstruction error.  
The red windows correspond to the anomalous windows labeled in the dataset.  
The yellow windows correspond to anomalous windows detected by 
\texttt{LPC-AD-SA}.  
From Figure \ref{fig:Imagescase}, 
one can observe that \texttt{LPC-AD-SA} identifies all anomalous windows out.  
The time series is well fitted by the reconstructions 
outside the anomalous windows, 
and is poorly fitted inside the anomalous windows.  

\noindent
{\bf Extensive empirical validation.}
We instantiate \texttt{LPC-Reconstruct} with the vanilla 
LSTM autoencoder, three simple predictors 
(i.e., linear predictor, LSTM enabled predictor and attention enabled predictor) 
and propose a simple randomized perturbation method based on Gaussian noise.  
We conduct extensive experiments on four public datasets that 
are widely used for evaluating MTS anomaly 
detection algorithms.  
Experiment findings are summarized into four folds.  
First, the \texttt{LPC-AD} reduces the training time of SOTA deep learning 
methods that focus on fast training by as high as 38.2\%.  
It improves the detection accuracy  of 
SOTA sophisticated deep learning based methods that 
focus on high accuracy by as high as 18.9\%. 
Second, \texttt{LPC-AD} has a high sample efficiency, 
i.e., reducing the training data points by 75\%, 
its detection accuracy only drops by less than 2\%.    
Third, \texttt{LPC-AD} is robust to its hyper parameters 
in terms of accuracy, i.e., the variation of detection accuracy is at most 
three percents when the hyper parameters varies in a fairly large range. 
Fourth, ablation study shows the necessity of each component 
and the benefit of each component.  
The highlights of our contributions include: 
\begin{itemize}
\item 
A generic architecture \texttt{LPC-Reconstruct}, 
which is a novel combination of ideas from autoencoder, 
predictive coding and randomized perturbation.  

\item 
A simple instantiation of \texttt{LPC-Reconstruct}, 
which contributes a new randomized perturbation method to avoid 
overfitting of anomalous dependence patterns.  

\item 
Extensive empirical evaluations to justify superior performance and 
reveal fundamental understating of our method.  
\end{itemize}

\section{\bf{Related works}}
\label{sec:relatedwork}

MTS anomaly detection is a challenging problem, and 
it has been studied extensively.  
From a methodological perspective, 
previous works can be categorized into two research lines, 
i.e., (1) classical methods and (2) deep learning based methods.  
Notable classical methods include 
\cite{bandaragoda2014efficient,kingsbury2020elle,liu2008isolation,patcha2007overview,wang2020real,yaacob2010arima}, 
just to name a few.  
One limitation of classical methods is that the expressiveness is low, 
resulting low anomaly detection accuracy \cite{xu2018unsupervised}.  
Deep learning based methods address this limitation.  
Our work falls into the research line of deep learning based methods.  
Methodologically, previous works on deep learning based 
MTS anomaly detection methods 
can be categorized into four groups: 
\textit{(1) forecasting based methods, (2) reconstruction based methods, 
(3) hybrid methods combining forecasting and reconstruction 
}, 
and 
\textit{ (4) miscellaneous methods. 
}

Forecasting based MTS anomaly detection methods  \cite{Bontemps2016,Chauhan2015,Deng2021,hundman2018detecting,Sakurada2014} 
use a prediction function to capture normal spatial and temporal 
dependence in MTS, 
and they identify abnormal dependence, i.e., report anomaly, 
when the prediction error is relatively large.    
The LSTM-AD \cite{Sakurada2014} uses a stacked LSTM to 
predict future data points from historical data points.  
It uses the likelihood of the prediction error to quantify the anomaly 
score of a data point, 
and it reports an anomaly whenever the anomaly score exceeds a given alert threshold.  
Chauhan \textit{et al.} applied the LSTM-AD to detect anomalies in electrocardiography time series \cite{Chauhan2015}. 
Bontemps \textit{et al.} \cite{Bontemps2016}  used the vanilla LSTM for forecasting  
and quantified the anomaly score via the prediction errors of several steps.  
LSTM-NDT\cite{hundman2018detecting} also uses the vanilla LSTM for forecasting and uses the prediction error of 
a data point to quantify the anomaly score.  
It develops a non-parameterized method to select the alert threshold.   
However, the detection accuracy of 
these LSTM based MTS anomaly detection methods 
\cite{Bontemps2016,Chauhan2015,Sakurada2014,hundman2018detecting} 
is not high enough, because LSTM alone lacks expressiveness to capture complex dependence in MTS \cite{tuli2022tranad}.  
Deng \textit{et al.} \cite{Deng2021} proposed 
a graph neural network to capture spatial dependence 
of multi-time series explicitly 
and used a graph attention network to predict future data points 
from historical data points.  
However, the graph attention network makes this method 
unable to capture long-term dependence well.  
These forecasting based methods have a relatively fast training speed,  
but their detection accuracy is greatly outperformed by reconstruction 
based methods \cite{li2021multivariate,su2019robust,tuli2022tranad}.  
 
Reconstruction based methods 
capture the normal spatial and temporal dependence in multi-time series 
via a low dimensional latent representation, 
and they identify abnormal dependence, i.e., report anomaly, 
when the reconstruction from the low dimensional latent representation differs from the actual observation.    
From a model architecture's perspective, 
previous works can be categorized into 
autoencoder based ones \cite{audibert2020usad,Li2021,Malhotra2016,tuli2022tranad,zhang2019deep,Huang2020} 
and variational autoencoder based ones \cite{dai2021sdfvae,li2021multivariate,park2018multimodal,su2019robust,xu2018unsupervised}.

\begin{itemize}
\item 
{\bf Autoencoder (AE) based methods.}  
The EncDec-AD \cite{Malhotra2016} 
instantiates the autoencoder architecture with the vanilla LSTM for MTS 
anomaly detection.  
The reconstruction error of a data point quantifies the anomaly score, 
and EncDec-AD reports an anomaly whenever the anomaly score exceeds a given threshold.  
The MSCRED \cite{zhang2019deep} improves the expressiveness of EncDec-AD.
It uses a convolutional network to capture spatial dependence and 
a convolutional LSTM to capture temporal dependence.   
However, it is time-consuming to train the model, and it requires 
a large amount of training data to fit the model well.   
Recent autoencoder based methods \cite{audibert2020usad,Li2021,tuli2022tranad}  
contribute simple models with a fast training speed, 
while retaining a fairly high detection accuracy.  
The openGauss \cite{Li2021} instantiates the autoencoder architecture with 
a tree-based LSTM for MTS anomaly detection. 
It has a fast training speed and low memory requirement.  
USAD \cite{audibert2020usad} further improves the 
training speed via adversarial training.  
It has one simple encoder and two decoders.  
Two decoders share the same encoder, and 
the purpose of two decoders is to enable adversarial training, 
which amplifies the temporal anomaly.  
The encoder and two decoders are composed of several linear layers, 
enabling a fast training speed.  
However, simple linear layers cannot capture temporal dependence well 
because they can not process longer input.   
TranAD \cite{tuli2022tranad} replaced the linear layer of USAD 
with transformer\cite{vaswani2017attention} network 
to capture temporal dependence better.  
TranAD contributes a slightly more sophisticated model than USAD. 
It is shown to have a higher detection accuracy than USAD, 
while retaining a fast training speed.  
Compared to these works, our work presents a new architecture, 
which combines autoencoder, 
predictive coding and randomized perturbation.  
Furthermore, our work outperforms them in both training speed 
and detection accuracy.  

\item {\bf Variational autoencoder (VAE) based methods.}:  
The Donut \cite{xu2018unsupervised} extends the vanilla VAE by 
changing the linear layers to represent the variance of latent variables to 
 a layer with the soft plus operation.   
The LSTM-VAE \cite{park2018multimodal} instantiates the VAE architecture by 
replacing the feed-forward network of VAE with a vanilla LSTM network to 
capture the temporal dependence in MTS.   
To improve the expressiveness of Donut and LSTM-VAE, 
OmniAnomaly \cite{su2019robust} changes the latent variables' distribution of 
LSTM-VAE from standard Gaussian distribution
to a more sophisticated distribution. 
The sophisticated distribution in OmniAnomaly is a combination of 
a linear Gaussian state space model \cite{kitagawa1996linear} 
and a normalizing flow \cite{rezende2015variational}.   
OmniAnomaly further uses latent variable linking to enhance the temporal dependence between latent variables.  
InterFusion \cite{li2021multivariate} improves OmniAnomaly by 
using an anomaly pre-filtering algorithm to filter out potential anomalous data in the training set.  
It also compresses the multi-time series 
via two-view embedding.  
To better capture spatial dependence in MTS, 
the SDFVAE \cite{dai2021sdfvae} 
instantiates the VAE architecture with a convolutional neural network, a BiLSTM, and recurrent VAE.    
Although OmniAnomaly, InterFusion and SDFVAE have high expressiveness 
because of their sophisticated models, 
their training processes are very resource-intensive and time-consuming.   
Different from them, our work is an autoencoder based method, 
and it outperforms them in both training speed 
and detection accuracy.  
\end{itemize}

Hybrid methods combine forecasting-based methods and reconstruction based methods 
in the sense that they jointly optimize a forecasting model and a reconstruction model.   
They attempt to overcome the shortcomings that forecasting based methods 
and reconstruction based methods suffer on their own \cite{Srivastava2015}.  
One notable architecture of hybrid methods is proposed by Srivastava \textit{et al.} \cite{Srivastava2015}
This architecture contains one encoder and two decoders sharing the same encoder.  
One of the decoders is designed for forecasting purposes, 
and the other is designed for reconstruction purposes.  
In the seminal work on this architecture, Srivastava \textit{et al.} \cite{Srivastava2015} 
instantiated this architecture with the vanilla LSTM.  
Medel \textit{et al.} \cite{Medel2016} instantiated this architecture 
with convolutional LSTM to have a high expressiveness capability.  
Zhao \textit{et al.} \cite{Zhao2017} instantiated this architecture 
with a 3D convolutional neural network 
for better detection of video series anomalies.  
These methods are mainly designed for video series, 
and they do not capture the spatial and temporal dependence in multi-time series well. 
The MTAD-GAT \cite{Zhao2020} is composed of a graph attention network for forecasting, 
which captures the spatial dependence,  
and a vanilla VAE for reconstruction, which captures the likelihood of a data point.  
It jointly optimizes those two models and quantifies the anomaly score with 
both the prediction error and the reconstruction error for anomaly detection.  
The CAEM \cite{Zhao2020} is composed of a convolutional autoencoder 
for reconstruction and a BiLSTM for forecasting.  
It also jointly optimizes those two models and quantifies the anomaly score with 
both the prediction error and the reconstruction error for anomaly detection.  
However, these methods have a high computational cost in training 
and low scalability for high-dimensional datasets.  
They are outperformed by autencoder based methods 
like TranAD \cite{tuli2022tranad}.   

Miscellaneous methods refer to notable related works that do not fall into the above research lines.  
The DAGMM \cite{zong2018deep} is composed of a simple autoencoder model for dimensionality reduction 
and a Gaussian Mixture Model \cite{reynolds2009gaussian} for density estimation.  
The density estimated by the GMM is used to predict the next data point.  
However, it is computationally expensive for training and unable to capture the spatial-temporal dependence well.    
Ren \textit{et al.} \cite{ren2020estimate} proposed estimators for the implicit likelihoods of GANs, 
and they showed that such estimators could be applied to detect anomalies 
in multi-time series data.    
The MAD-GAN \cite{li2019mad} instantiated the vanilla GAN architecture with LSTM-RNN.  
Koopman operator is a classical method in control theory \cite{lusch2018deep}. 
It has a similar idea of latent predictive coding. 
Theoretically, in an infinite-dimensional latent state space, 
the next state can be a linear function of the states in a recent time window \cite{lusch2018deep}.  
In particular, when we instantiate the predictor in our LPC-AD architecture as a linear function,  
we get a deep Koopman operator \cite{lusch2018deep}.
For practical applications where we cannot have finite-dimensional states, 
non-linear predictor function is needed for state transitions in latent space.  
This is validated by our experiment in Figure \ref{fig:Ablation Study}, 
where we observe that using non-linear predictor functions 
improves the accuracy of anomaly detection.

Last, in terms of design objective, our work is closely related to USAD \cite{audibert2020usad} and TranAD \cite{tuli2022tranad}.  
To the best of our knowledge, USAD and TranAD are the only two works that 
aim to reduce training time while retaining a fairly high detection accuracy.  
We proposed a generic architecture and show that simple instantiations of this 
architecture can outperform USAD and TranAD in training speed 
and outperform SOAT methods (including sophisticated models) in detection accuracy.   
The design of our architecture reveals new insights in exploring latent predictive coding 
and randomized perturbation for MTS anomaly detection.


\section{\bf{Problem Formulation} }

We consider an $M \in \mathbb{N}_+$ dimensional time series 
indexed by time stamps $t \in \mathbb{N}_+$.  
Let $\bm{x}_t \triangleq (x_{t,1}, \ldots, x_{t,M}) \in \mathbb{R}^M$ denote the $t$-th 
observation or sample point of the time series.     
For example, $\bm{x}_t$ can be the values of $M$ KPIs 
of a cloud computing system measured at time stamp $t$.   
Note that in practice, the observations of a time series are sampled 
at a fixed rate or time varying rate.  
In this paper, we do not make any assumptions about the sampling rate.  
We consider the setting that we are given a training dataset 
of $T \in \mathbb{N}_+$ data points of the time series 
denoted by $\mathcal{T} = \{ \bm{x}_1, \ldots, \bm{x}_T \}$.  
Each data point is not associated with a label on whether it is anomalous or not.  
Our objective is to design and train an anomaly detection 
algorithm from the training dataset $\mathcal{T}$.

To make our model more robust, we normalize the time series data as follows:
\begin{equation}
x_{t,m} 
\leftarrow 
\frac{ x_{t,m} -  \min_{\tau \in \{1,\ldots, T\}} x_{\tau,m} }
{ \max_{\tau \in \{1,\ldots, T\}} x_{\tau,m} - \min_{\tau \in \{1,\ldots, T\}} x_{\tau,m} + \alpha }, 
\quad
\forall t, m,
\end{equation}
where $\alpha \in \mathbb{R}_+$ denotes a smoothing factor.   
The smoothing factor $\alpha$ is a small positive constant vector, 
which prevents zero-division.  

\section{The LPC-AD Algorithm}

Our algorithm is built on two sliding windows, i.e., a sliding window of historical data points and a sliding window of future data points.
It utilizes the dependence between data points in these two sliding windows to detect the anomaly.  
Formally, denote a sliding window of $\ell_h \in \mathbb{N}_+$ latest historical data points up to time stamp $t$ as 
\begin{align}
{\bm W}^h_t \triangleq [ {\bm x}_{t- \ell_h+1},\cdots, {\bm x}_t ],   
\end{align}
where $t \geq \ell_h$. 
Denote a sliding window of $\ell \in \mathbb{N}_+$ future data points 
starting from time stamp $t+1$ as 
\begin{equation}
\bm{W}_{t+1} = [ \bm{x}_{t+1}, \ldots, \bm{x}_{t+\ell}].  
\end{equation}
The window size $\ell_h$ and $\ell$ are two hyperparameters, 
and in general, they may be different, i.e., $\ell_h \neq \ell$.  
We aim to design and train an algorithm to learn the dependence between 
two consecutive windows, i.e., ${\bm W}^h_t$ and $\bm{W}_{t+1}$, from $\mathcal{T}$.  
And then, we utilize it to decide whether a data point $\bm{x}_t$ outside the training dataset, 
where $t > T$, is an anomaly or not.

Formally, we design an anomaly detection algorithm based on the latent predictive coding. 
Algorithm \ref{alg:LPC-AD} outlines our LPC-AD 
(\underline{L}atent \underline{P}redictive \underline{C}oding for \underline{A}nomaly \underline{D}etection) algorithm.  
The LPC-AD takes two consecutive sliding windows, 
i.e., ${\bm W}^h_t$ and ${\bm W}_{t+1}$, 
and an alert threshold $\lambda \in \mathbb{R}_+$ as input.  
Note that the sliding window ${\bm W}_{t+1}$ is outside the training dataset, i.e., $t\geq T$.  
It outputs an $\ell$-dimensional binary vector 
to indicate where a data point in the window ${\bm W}_{t+1}$ is anomalous or not, 
In Step \ref{LPC-AD:noise} to \ref{LPC-AD:score}, 
an anomaly score for each data point in ${\bm W}_{t+1}$ is computed, 
which quantifies the likelihood for a data point to be an anomaly.   
More specifically, the anomaly score is built on two two key ideas.  
The one is to utilize latent predictive coding to 
capture the normal dependence between two consecutive sliding windows.  
This is achieved by the $\texttt{LPC-Reconstruct}( \bm{W}^h_t, \bm{W}_{t+1}, \bm{\epsilon} ; \bm{\Theta})$ 
algorithm in Step \ref{LPC-AD-reconstruct}. 
The details of $\texttt{LPC-Reconstruct}$ are deferred to Section \ref{sec:LPC-reconstruct}.  
The other idea is random perturbation, which is 
achieved by generating a stochastic noise in Step \ref{LPC-AD:noise} 
and taking it as an input to the 
$\texttt{LPC-Reconstruct}$
algorithm in Step \ref{LPC-AD-reconstruct}.   
The purpose of this random perturbation 
is to avoid overfitting the patterns of anomaly data points.  
Based on the anomaly score, in the remaining steps, 
we use the alert threshold $\lambda$ to report the anomaly.  
In particular, when the anomaly score exceeds the alert threshold $\lambda$, 
we report an anomaly; otherwise, we report no anomaly.

\begin{algorithm}[htb]
	\caption{LPC-AD Algorithm}
	\label{alg:LPC-AD}
	\begin{algorithmic}[1]
	\Require
	$({\bm W}^h_t, {\bm W}_{t+1}) , \forall t\geq T$, alert threshold $\lambda$
	\Ensure $\bm{a} \triangleq (a_1, \ldots, a_\ell)$
	\State $\bm{\epsilon} \sim \mathcal{N} (\bm{0}, \bm{\Sigma})$ 
	\label{LPC-AD:noise}
	\State $[\widetilde{\bm{x}}_{t+1}, \ldots,\widetilde{\bm{x}}_{t+\ell} ] 
	\leftarrow \texttt{LPC-Reconstruct}( \bm{W}^h_t, \bm{W}_{t+1}, \bm{\epsilon} ; \bm{\Theta})$
	\label{LPC-AD-reconstruct}
	\State $\text{AnomalyScore}_i \leftarrow \Vert \bm{x}_{t+i} - \widetilde{\bm{x}}_{t+i} \Vert_2, \forall i=1,\ldots, \ell$
	\label{LPC-AD:score}
	\State $a_{i} \leftarrow \mathbb{I}_{\{\text{AnomalyScore}_i \geq \lambda \}}, \forall i = 1,\ldots, \ell$.
	\State {\bf Return} $\bm{a}$. 
	\end{algorithmic}
\end{algorithm}
 
\noindent
{\bf Remark 1: selection of $\lambda$.} 
Using an alert threshold to report anomaly is 
widely used in previous works \cite{audibert2020usad,tuli2022tranad,li2021multivariate,reynolds2009gaussian,su2019robust}.  
Though few adaptive methods were proposed to select the alert threshold automatically, 
these methods turned out to be either not stable or computationally expensive.  
As a consequence, exhaustive search method is the most widely adopted method 
\cite{audibert2020usad,tuli2022tranad,li2021multivariate,reynolds2009gaussian,su2019robust}.  
The exhaustive search method only requires proper discretization of the 
domain of the alert threshold.  
The domain of the alert threshold depends on the underlying anomaly score evaluation method, 
which can be quite different across different methods 
\cite{audibert2020usad,tuli2022tranad,li2021multivariate,reynolds2009gaussian,su2019robust}.

\noindent
{\bf Remark 2: multiple windows.} 
When there are multiple sliding window pairs to detect, 
i.e., $\{ ({\bm W}^h_{t_i}, {\bm W}_{t_i+1}) : i = 1, \ldots, I \}$, 
where $I \in \mathbb{N}_+$,  
one can apply Algorithm \ref{alg:LPC-AD} to each pair of them sequentially or in parallel.  


\section{Design and Training of \texttt{LPC-Reconstruct}}
\label{sec:LPC-reconstruct}

In this section, we first present the general architecture of the  
$\texttt{LPC-Reconstruct}( \bm{W}^h_t, \bm{W}_{t+1}, \bm{\epsilon} ; \bm{\Theta})$ 
algorithm.  
Then, we present a simple instantiation of the general architecture 
to demonstrate the power of the proposed architecture.  
Lastly, we design an algorithm to train this instantiation 
from the training dataset $\mathcal{T}$.

\subsection{General Model Architecture}

\begin{figure}[htb]
	\centering 
	\includegraphics[width=0.6\textwidth]{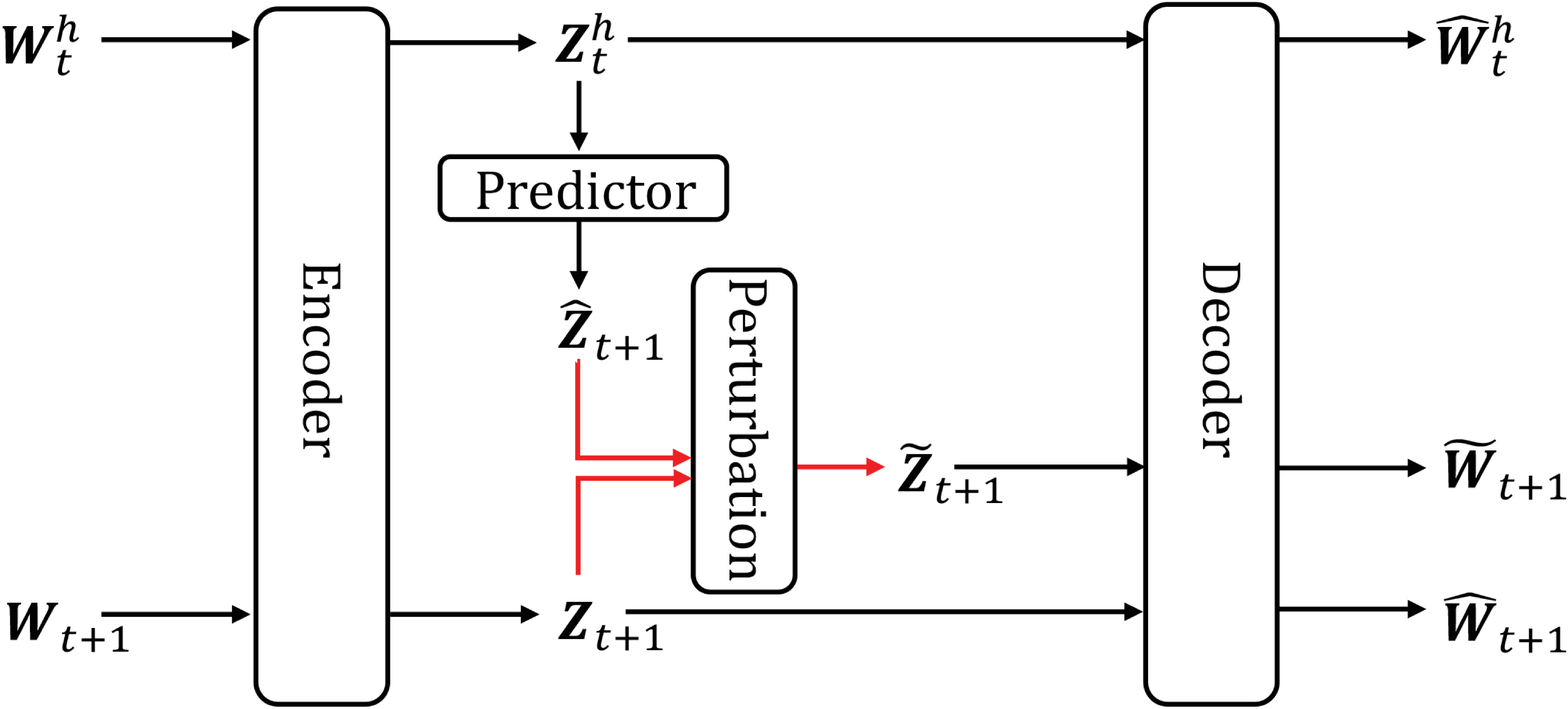}
	\caption{Architecture of $\texttt{LPC-Reconstruct}( \bm{W}^h_t, \bm{W}_{t+1}, \bm{\epsilon} ; \bm{\Theta})$.} 
	\label{fig:Model Architecture}
\end{figure}

We parameterize $\texttt{LPC-Reconstruct}( \bm{W}^h_t, \bm{W}_{t+1}, \bm{\epsilon} ; \bm{\Theta})$  
via a neural network as shown in   
Figure \ref{fig:Model Architecture}.  
Four building blocks of this architecture are: 
a sequence encoder 
$
\texttt{SeqEnc}(\cdot; \bm{\Theta}_{\text{SE}}) 
$
with parameter $\bm{\Theta}_{\text{SE}}$,  
a sequence decoder 
$
\texttt{SeqDec}(\cdot; \bm{\Theta}_{\text{SD}}) 
$  
with parameter $\bm{\Theta}_{\text{SD}}$, 
a predictor 
$
\texttt{Predic}(\cdot; \bm{\Theta}_{\text{PD}}) 
$
with parameter $\bm{\Theta}_{\text{PD}}$
and a randomized data perturbation operator  
$
\texttt{RandPerturb}(\cdot; \bm{\Theta}_{\text{RP}})
$
with parameter $\bm{\Theta}_{\text{RP}}$.  
We defer the design details of each of these building blocks to Section \ref{sec:LPC-Recons-Instance}, 
and let us state some properties of them first, 
which are helpful for us to deliver the key ideas of our model architecture. 

\begin{itemize}
\item 
$\texttt{SeqEnc}(\cdot ; \bm{\Theta}_{\text{SE}})$.  
It takes a consecutive sequence of data points $\bm{x}_{t_1}, \bm{x}_{t_1 + 1}, \ldots, \bm{x}_{t_2}$ 
as the input, where $t_1, t_2 \in \{1, \ldots, T\}$ 
and $t_1 \leq t_2$, 
and outputs an encoded low dimensional data sequence denoted by 
$\bm{z}_{t_1}, \bm{z}_{t_1 + 1}, \ldots, \bm{z}_{t_2}$, 
where $\bm{z}_t \in \mathbb{R}^N$, $t \in \{t_1, \ldots, t_2\}$, $N \in \mathbb{N}_+$ and $N < M$.   
We call $\bm{z}_t$ the latent variable and it is 
a compression of the original data $\bm{x}_{t}$.  

\item 
$\texttt{SeqDec}(\cdot; \bm{\Theta}_{\text{SD}})$.  
It takes a low dimensional latent variable sequence $\bm{z}_{t_1}, \ldots, \bm{z}_{t_2}$ as the input, 
and outputs a high dimensional decoded sequence denoted by 
$\widehat{\bm{x}}_{t_1}, \ldots, \widehat{\bm{x}}_{t_2}$, 
where $\widehat{\bm{x}}_t \in \mathbb{R}^M$, $t \in \{t_1, \ldots, t_2\}$.   
The $\widehat{\bm{x}}_{t}$ is a reconstruction of the original data $\bm{x}_{t}$.  

\item 
$\texttt{Predic}(\cdot; \bm{\Theta}_{\text{PD}})$.  
It takes a consecutive sequence of $\ell_h$ latent variables 
$\bm{z}_{t-\ell_t+1}, \ldots, \bm{z}_{t}$
as the input, 
and outputs a prediction on the next $\ell$ latent variables 
$\bm{z}_{t+1}, \ldots, \bm{z}_{t+\ell}$.   

\item 
$\texttt{RandPerturb}(\cdot, \bm{\epsilon}; \bm{\Theta}_{\text{RP}})$.  
It takes the latent variable sequence predicted by $\texttt{Predic}(\cdot; \bm{\Theta}_{\text{PD}})$ and the ground truth 
latent variable sequence as the input, and generates a zero mean stochastic noise vector $\bm{\epsilon}$ to perturb the predictions.  
\end{itemize}

In $\texttt{LPC-Reconstruct}$, we first input the data points of two consecutive sliding window, 
i.e., ${\bm W}^h_t, {\bm W}_{t+1}$, into the encoder to obtain a low dimensional representation 
${\bm Z}^h_t, {\bm Z}_{t+1}$, where 
\[
[{\bm Z}^h_t, {\bm Z}_{t+1}] = \texttt{SeqEnc}({\bm W}^h_t, {\bm W}_{t+1}; \bm{\Theta}_{\text{SE}}),  
\]
and ${\bm Z}^h_t = [ {\bm z}_{t- \ell_h1+1},\cdots, {\bm z}_t ]$ 
and ${\bm Z}^2_t = [ {\bm z}_{t+1},\cdots, {\bm z}_{t+ \ell} ]$.  
The purpose of this low dimensional representation is to eliminate 
the redundancy in the time series data.  
Then we use the decoder to reconstruct the original time series from 
the encoded low dimensional data, i.e., ${\bm Z}^h_t$ and ${\bm Z}_{t+1}$, 
\[
[ \widehat{\bm{W}}^h_t, \widehat{\bm{W}}_{t+1}] 
= 
\texttt{SeqDec}({\bm Z}^h_t, {\bm Z}_{t+1}; \bm{\Theta}_{\text{SD}}), 
\] 
where $\widehat{\bm{W}}^h_t = [ \widehat{\bm{x}}_{t- \ell_h+1},\cdots, \widehat{\bm{x}}_t ]$ 
and $\widehat{\bm{W}}_{t+1} = [ \widehat{\bm{x}}_{t+1},\cdots, \widehat{\bm{x}}_{t+ \ell} ]$.  
To capture the dependence in the low dimensional representation, 
we use a prediction operator, 
which use the encoded data of the historical window to predict the data associated with the future window, formally   
\[
\widehat{{\bm Z}}_{t+1} 
= 
\texttt{Predic}({\bm Z}^h_t; \bm{\Theta}_{\text{PD}}).  
\]
Then we use a randomized perturbation operator to perturb the above prediction. 
\[
\widetilde{\bm{Z}}_{t+1} 
=
\texttt{RandPerturb}({\bm Z}_{t+1}, \widehat{{\bm Z}}_{t+1}, \bm{\epsilon}; \bm{\Theta}_{\text{RP}}), 
\]
where $\widetilde{\bm{Z}}_{t+1} =  [ \widetilde{\bm{z}}_{t+1},\cdots, \widetilde{\bm{z}}_{t+ \ell} ]$ 
and $\bm{\epsilon}$ denotes an $\ell$-dimensional zero mean stochastic noise.  
The purpose of this perturbation is to avoid overfitting  
the patterns of anomaly data points, 
as there may be some anomaly data points in the training data.   
This operation also makes $\widetilde{\bm{Z}}_{t+1}$ more robust to small outliers in training data.  
It has a theoretical foundation 
in online learning in the presence of adversarial data \cite{Abernethy2016}.  
Finally, we obtain the reconstructed data as:  
\[
[ \widetilde{\bm{W}}^h_t, \widetilde{\bm{W}}_{t+1}] 
= 
\texttt{SeqDec}({\bm Z}^h_t, \widetilde{\bm{Z}}_{t+1}; \bm{\Theta}_{\text{SD}}), 
\]
where $\widetilde{\bm{W}}_{t+1}$ serves as the output of 
$
\texttt{LPC-Reconstruct}( \bm{W}^h_t, \bm{W}_{t+1}, \bm{\epsilon} ; \bm{\Theta}): 
$ 
\[
\texttt{LPC-Reconstruct}( \bm{W}^h_t, \bm{W}_{t+1}, \bm{\epsilon} ; \bm{\Theta}) 
= 
\widetilde{\bm{W}}_{t+1}.  
\]
Then, the parameter $\bm{\Theta}$ of 
$
\texttt{LPC-Reconstruct}( \bm{W}^h_t, \bm{W}_{t+1}, \bm{\epsilon} ; \bm{\Theta}) 
$
can be summarized as:
\[
\bm{\Theta} = [\bm{\Theta}_{\text{SE}}, \bm{\Theta}_{\text{SD}}, \bm{\Theta}_{\text{PD}}, \bm{\Theta}_{\text{RP}}].  
\]

To learn the function $\texttt{LPC-Reconstruct}( \bm{W}^h_t, \bm{W}_{t+1}, \bm{\epsilon} ; \bm{\Theta})$ from the 
training data $\mathcal{T}$, we consider a loss function associated with time stamp $t$ defined as: 
\[
L_t (\bm{\Theta})
= 
\underbrace{ 
\Vert {\bm W}^h_t - {\widehat{\bm W}}^h_t\Vert_2 + \Vert {\bm W}_{t+1} - {\widehat{\bm W}}_{t+1}\Vert_2 
}_{
\text{redundancy reduction loss}
}
+
\underbrace{ 
\mathbb{E}_{\bm{\epsilon}} [ \Vert  \bm{W}_{t+1}  - \widetilde{\bm{W}}_{t+1} \Vert_2 ]
}_{
\text{dependency capturing loss}
},
\]
where the expectation notation $\mathbb{E}_{\bm{\epsilon}}$ 
means take expectation with respect to the random vector $\bm{\epsilon}$.  
Given a training dataset $\mathcal{T}$, the total loss is 
\[
L(\bm{\Theta}) 
=
\sum\nolimits_{t=\ell_h}^{T-\ell} L_t (\bm{\Theta}).  
\]
The physical meaning of the above loss is to find the parameters $\bm{\Theta}$ 
that best capture the dependence between consecutive sliding windows.  
Note the in the above loss, we start from $t=\ell_h $  
instead of $t=1$, 
to avoid the corner case where the historical window $\bm{W}^h_t$ 
contains less than $\ell_h$ data points.  
Similarly, we end at $t = T-\ell-1$ instead of $t=T$, 
to avoid the corner case that the future window $\bm{W}_{t+1}$ 
contains less than $\ell$ datapints.  

\noindent
{\bf Remark.} 
The architecture of $\texttt{LPC-Reconstruct}( \bm{W}^h_t, \bm{W}_{t+1}, \bm{\epsilon}; \bm{\Theta})$  
depicted in Figure \ref{fig:Model Architecture} is quite general.  
This architecture allows one to select different instantiations of the architecture 
to attain different tradeoffs between detection accuracy and computational cos.

\subsection{Instantiation on the Model} 
\label{sec:LPC-Recons-Instance}

To demonstrate the power of our proposed architecture depicted in Figure \ref{fig:Model Architecture}, 
we present simple instantiations of it.  
As one will see in Section \ref{sec:Experiment}, 
these simple instantiations can already outperform SOTA algorithms 
in terms of both the detection accuracy and training speed.   

\begin{figure}[htb]
	\centering 
	\subfigure[Architecture of $\texttt{SeqEnc}(\cdot ; \bm{\Theta}_{\text{SE}})$]{
		\centering
		\includegraphics[width=0.4\textwidth]{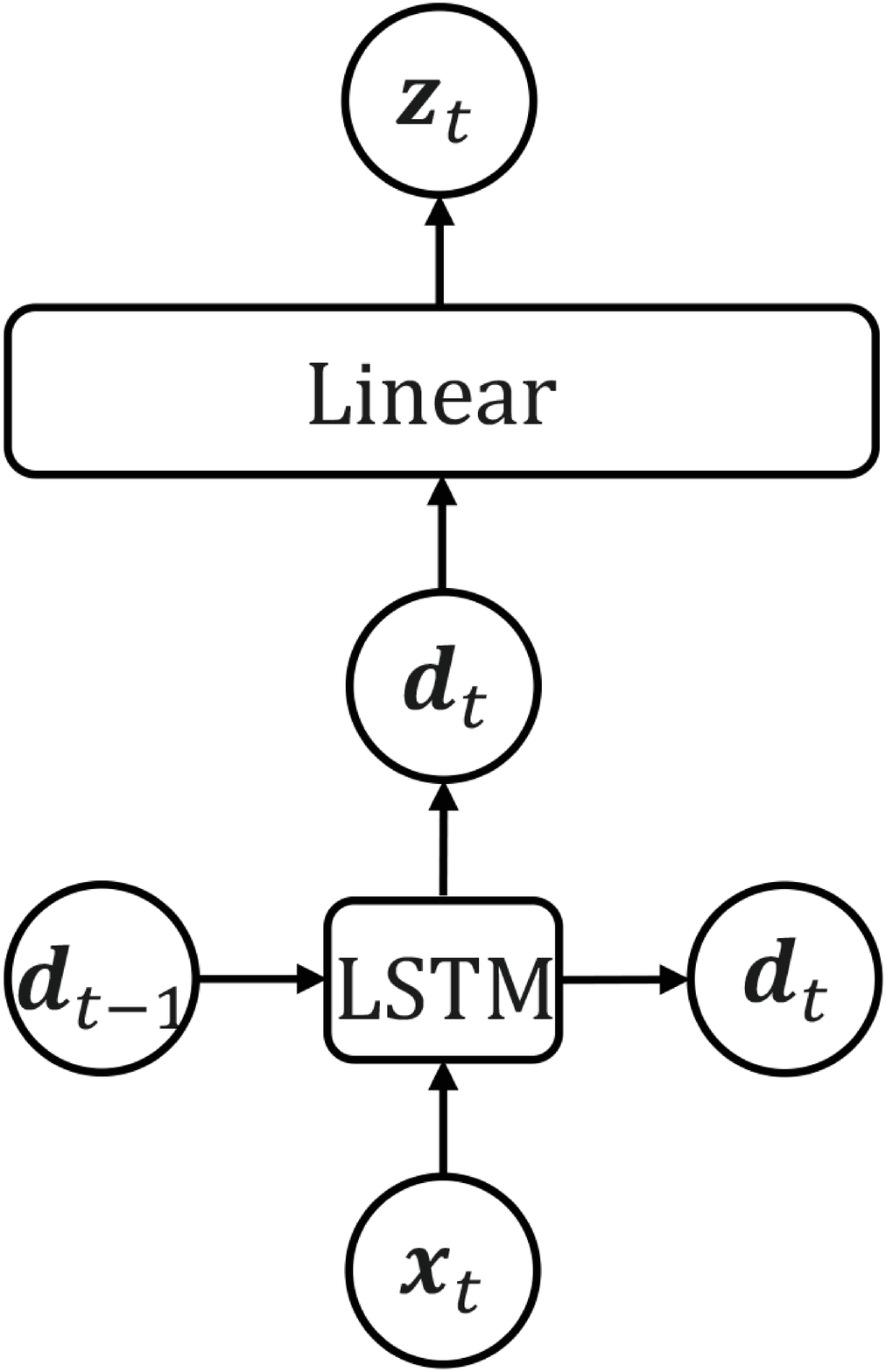}
		\label{fig:Encoder}
	}
	\subfigure[Architecture of $\texttt{SeqDec}(\cdot; \bm{\Theta}_{\text{SD}})$]{
		\centering
		\includegraphics[width=0.4\textwidth]{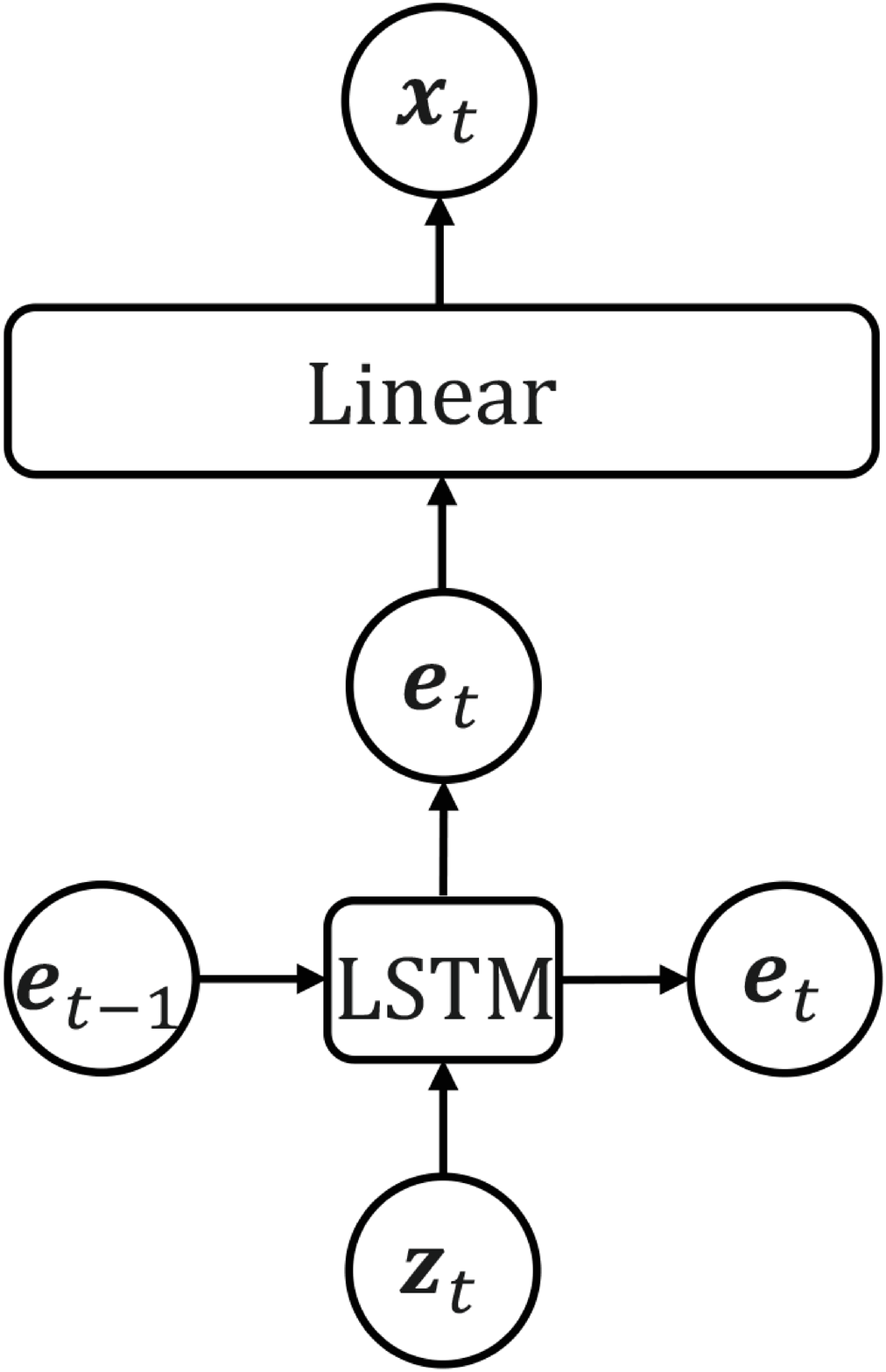}
		\label{fig:Decoder}
	}
	\caption{Instantiate $\texttt{SeqEnc}(\cdot ; \bm{\Theta}_{\text{SE}})$ and $\texttt{SeqDec}(\cdot; \bm{\Theta}_{\text{SD}})$.  
	} 
	\label{fig:Submodel Architecture} 
\end{figure}

\noindent
{\bf Instantiation of } $\texttt{SeqEnc}(\cdot ; \bm{\Theta}_{\text{SE}})$ 
and $\texttt{SeqDec}(\cdot; \bm{\Theta}_{\text{SD}})$.  
We instantiate the sequence encoder $\texttt{SeqEnc}(\cdot ; \bm{\Theta}_{\text{SE}})$ 
and the sequence decoder $\texttt{SeqDec}(\cdot; \bm{\Theta}_{\text{SD}})$ 
by adding a linear layer to the classical LSTM.  
The architectures of them are shown in Figure \ref{fig:Encoder} and Figure \ref{fig:Decoder}.  
The purpose of adding a linear layer to the classical LSTM is to 
reduce the number of dimensions and eliminate redundant information.  
In particular, a more sophisticated sequence encoder 
and sequence decoder can have better performance in dimension reduction 
which may, in turn, improve the final detection accuracy of the architecture.  
Since we are processing a multi-variate time series data 
with complex temporal dependency, 
a simple encoder composed of linear layers is oversimplified to encode the original data well \cite{audibert2020usad}.  
Recurrent neural networks are more sophisticated than 
this oversimplified linear encoder, 
and they can capture temporal dependency in the time series data.  
However, simple RNNs suffer from the gradient vanishing issue, 
which makes them unsuitable to capture long dependence in the data.  
One simple variant of RNNs, which is LSTM, can solve 
this gradient vanishing issue. 

\noindent
{\bf Instantiation of} $\texttt{Predic}(\cdot; \bm{\Theta}_{\text{PD}})$. 
Again we consider simple instantiations of the predictor $\texttt{Predic}(\cdot; \bm{\Theta}_{\text{PD}})$, 
to better demonstrate the power of our proposed architecture.  
We consider three instances of $\texttt{Predic}(\cdot; \bm{\Theta}_{\text{PD}})$. 
\begin{enumerate}
\item 
{\bf Instantiating $\texttt{Predic}(\cdot; \bm{\Theta}_{\text{PD}})$ using linear transformation. 
}  
This is the simplest instance of $\texttt{Predic}(\cdot; \bm{\Theta}_{\text{PD}})$.  
This instance enables us to compare our architecture 
with the methods based on the Koopman operator theory \cite{lusch2018deep}, 
which uses a linear transformation to capture dependence between latent embeddings.  
Formally, for a latent embedding sequence 
${\bm Z}_t = [{\bm z}_{t-\ell_h+1},\cdots,{\bm z}_t]$, where ${\bm z}_t$ is a $N$-dimensional vector, 
the linear transformation predicts 
the latent embedding vector in the next time slot as 
\begin{equation}
\texttt{Predic}({\bm Z}^h_t; \bm{\Theta}_{\text{PD}})  
= {\bm P}{\bm Z}_t\bm Q, 
\label{Predictor:LinearTransfor}
\end{equation}
where parameters ${\bm P} \in \mathbb{R}^{N\times N}$, $\bm{Z}_t \in \mathbb{R}^{N\times \ell_h}$ and ${\bm Q} \in \mathbb{R}^{\ell_h}$.

\item 
{\bf Instantiating $\texttt{Predic}(\cdot; \bm{\Theta}_{\text{PD}})$ via LSTM enabled {\bf Seq2Seq}.  
}  
We consider a simple nonlinear instantiation of the predictor.  
This instantiation enables us to understand the impact of nonlinearity on the final 
detection accuracy of our proposed architecture.  
Formally, we instantiate $\texttt{Predic}(\cdot; \bm{\Theta}_{\text{PD}})$ with 
a basic seq2seq neural network complemented with a single LSTM layer.  
Figure \ref{fig:seq2seq} depicts details of this instantiation.  

\begin{figure}[htb]
	\centering
	\includegraphics[width=0.625\textwidth]{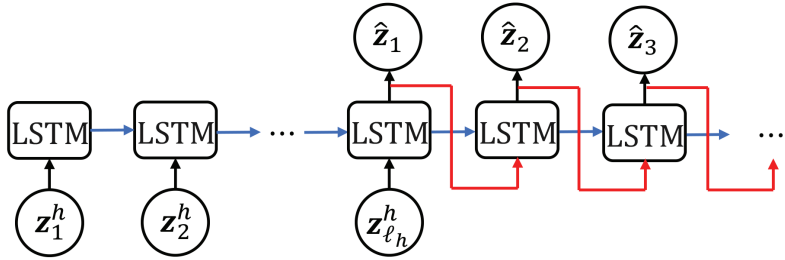}
	\caption{Instantiate $\texttt{Predic}(\cdot; \bm{\Theta}_{\text{PD}})$ via LSTM {\bf Seq2Seq}.} 
	\label{fig:seq2seq}
	\vspace{-0.08in}
\end{figure}

\item 
{\bf Instantiating $\texttt{Predic}(\cdot; \bm{\Theta}_{\text{PD}})$ via attention enabled {\bf Seq2Seq}. 
}  
This instance of $\texttt{Predic}(\cdot; \bm{\Theta}_{\text{PD}})$ is a little bit 
more sophisticated than the one depicted in Figure \ref{fig:seq2seq}.  
It was shown that seq2seq with attention mechanism 
is good at capturing temporal dependency and can achieve more accurate prediction \cite{qin2017dual}.   
Hence, the purpose of this instance is to understand the impact of 
the prediction accuracy of $\texttt{Predic}$ on the accuracy of anomaly detectiono in our \texttt{LPC-AD} framework.   
Formally, this instance of $\texttt{Predic}(\cdot; \bm{\Theta}_{\text{PD}})$ is developed in \cite{qin2017dual}.  
Figure \ref{fig:seq2seq-att} shows details on its architecture.
\begin{figure}[htb]
	\centering
	\vspace{-0.16 in}
	\includegraphics[width=0.56\textwidth]{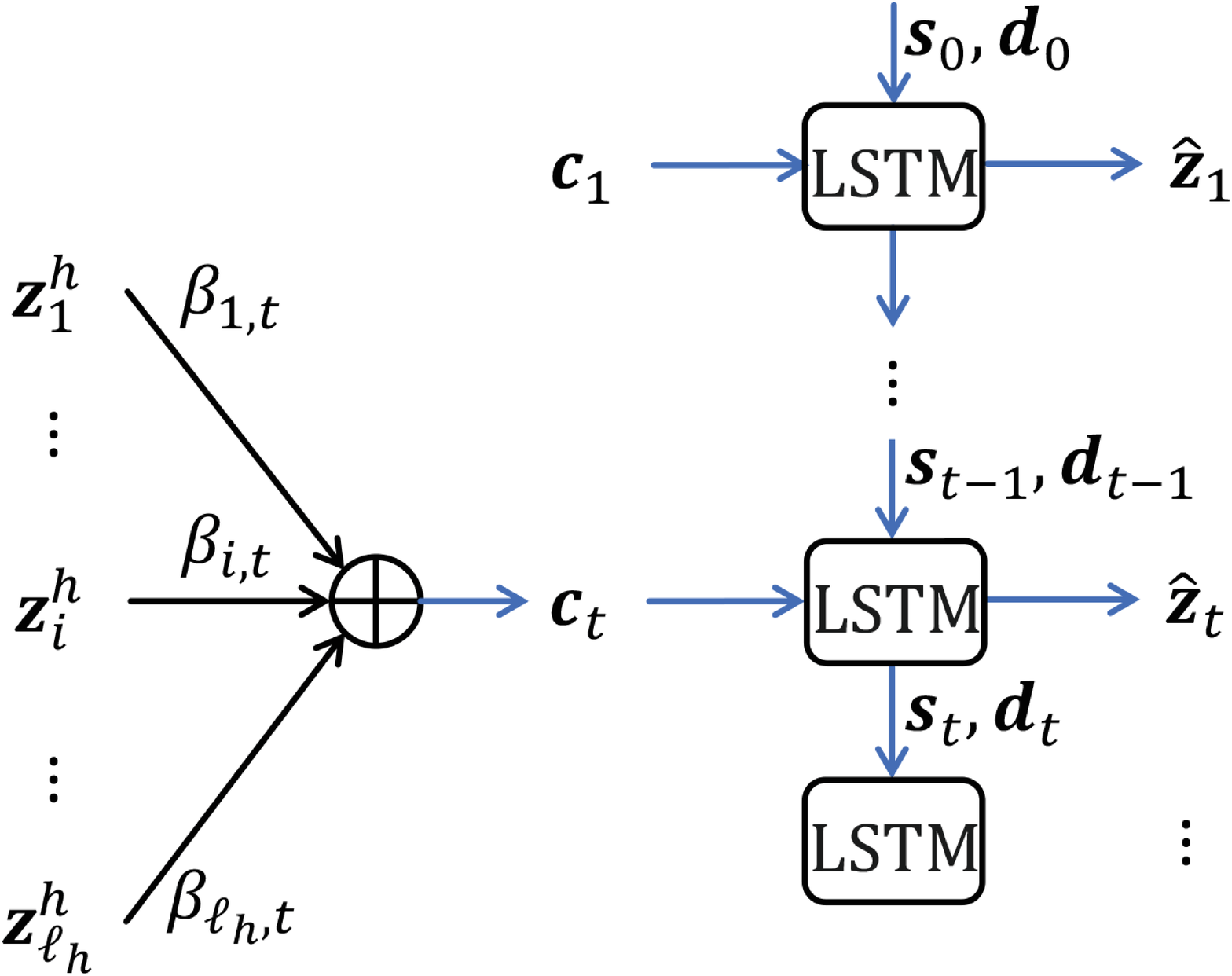}
	\caption{Instantiate $\texttt{Predic}(\cdot; \bm{\Theta}_{\text{PD}})$ via attention {\bf Seq2Seq}.} 
	\label{fig:seq2seq-att}
	\vspace{-0.16in}
\end{figure}

In Figure \ref{fig:seq2seq-att}, 
the  $[\beta_{1,t},\cdots,\beta_{\ell_h,t}]$ is the attention score vector, 
which is calculated by the attention function proposed by \cite{bahdanau2014neural}.  
Furthermore, we calculate the context vector ${\bm c}_t$ corresponding to predicted vector $\hat{\bm z}_t$ as follows: 
\begin{equation}
{\bm c}_t = \beta_{1,t} \times {\bm z}^h_1 + \cdots + \beta_{\ell_h, t} \times {\bm z}^h_{\ell_h},
\end{equation}
where $
\beta_{i,t} {=} \exp{l_{i,t}} / (\sum^{\ell_h}_{i=1}{\exp{l_{i,t}}}),
$
$
l_{i,t} {=} {\bm v} \tanh ({\bm W}[{\bm s}_{t-1}, {\bm d}_{d-1}] + {\bm U}{\bm z}^h_i),   
$
and the ${\bm v}$, ${\bm W}$ and ${\bm U}$ 
are the parameters of the attention function.
The attention enabled Seq2Seq can be trained in an end-to-end manner.

\end{enumerate}

\noindent
{\bf Instantiation of randomized perturbation} $\texttt{RandPerturb}(\cdot, \bm{\epsilon}; \bm{\Theta}_{\text{RP}})$.  
Again we consider simple instantiations on 
$\texttt{RandPerturb}(\cdot, \bm{\epsilon}; \bm{\Theta}_{\text{RP}})$, 
to better demonstrate the power of our proposed architecture.  
Note that $\bm{\epsilon}$ is a zero mean random vector generated from 
the Gaussian distribution $\mathcal{N} (\bm{0}, \bm{\Sigma})$.  
Formally, we use the following formula to instantiate 
$\texttt{RandPerturb}(\cdot, \bm{\epsilon}; \bm{\Theta}_{\text{RP}})$:
\[
\texttt{RandPerturb}({\bm Z}_{t+1}, \widehat{{\bm Z}}_{t+1}, \bm{\epsilon}; \bm{\Theta}_{\text{RP}}) 
= 
{\bm Z}_{t+1} + \bm{\epsilon} \odot ({\bm Z}_{t+1} - \widehat{{\bm Z}}_{t+1})^{\text{abs}}, 
\]
where $\odot$ denotes component-wise multiplication of vectors 
and $({\bm Z}_{t+1} - \widehat{{\bm Z}}_{t+1})^{\text{abs}}$ 
means taking the absolute of each component of the vector 
$
{\bm Z}_{t+1} - \widehat{{\bm Z}}_{t+1}. 
$

\subsection{Offline Training}

We observe that in the loss function $L_t (\bm{\Theta})$, 
the expectation 
$\mathbb{E}_{\bm{\epsilon}} [ \Vert  \widetilde{\bm{W}}_{t+1} - \bm{W}_{t+1} \Vert_2 ]$ 
is computationally expensive to evaluate.  
We use Monte Carlo simulation to address this 
computational issue.  
More specifically, we generate $K \in \mathbb{N}_+$ samples of 
the perturbation noise denoted by $\bm{\epsilon}_1, \ldots, \bm{\epsilon}_K$.  
Let ${\widetilde{\bm W}}_{t+1,k}$ denote the reconstructed data associated with $\bm{\epsilon}_k$.  
Formally, we approximate the expectation 
$\mathbb{E}_{\bm{\epsilon}} [ \Vert  \widetilde{\bm{W}}_{t+1} - \bm{W}_{t+1} \Vert_2 ]$ 
via  
$
\mathbb{E}[\Vert \widetilde{\bm W}_{t+1} - \bm{W}_{t+1} \Vert_2] 
\approx 
\frac{1}{K} \sum\nolimits^K_{k=1} \Vert {\widetilde{\bm W}}_{t+1,k} - {\bm W}_{t+1}\Vert_2.  
$
Finally, we use the following approximation on the per time slot 
loss to train the model:  
\begin{equation}
\label{equ:Loss}
L_t (\bm{\Theta})
{\approx}
\Vert {\bm W}^h_t {-} {\widehat{\bm W}}^h_t\Vert_2 
{+} \Vert {\bm W}_{t+1} {-} {\widehat{\bm W}}_{t+1}\Vert_2
{+}
\frac{1}{K} \sum^K_{k=1} \!\! \Vert {\widetilde{\bm W}}_{t+1,k} {-} {\bm W}_{t+1}\Vert_2.  
\end{equation}
Based on the above approximated loss, Algorithm \ref{alg: Training} outlines 
our model training procedure.  

\begin{algorithm}[htb]
	\caption{The Generic Training algorithm of LPC-AD}
	\label {alg: Training}
	\begin{algorithmic}[1]
	\Require
	$\{({\bm W}^h_{\ell_h}, {\bm W}_{\ell_h + 1}),({\bm W}^h_{\ell_h+1}, {\bm W}_{\ell_h+2}), \cdots, ({\bm W}^h_{T-\ell}, {\bm W}_{T-\ell+1})\}$, 
	number of training rounds $MaxEpoch$. 
	
	\Ensure
	Trained $\bm{\Theta}_{\text{SE}},\bm{\Theta}_{\text{SD}}, 
	\bm{\Theta}_{\text{RP}}, \bm{\Theta}_{\text{PD}}$
	\State Initialize $\bm{\Theta}_{\text{SE}},\bm{\Theta}_{\text{SD}}, 
	\bm{\Theta}_{\text{RP}}, \bm{\Theta}_{\text{PD}}$
	\State $e\leftarrow 1$
	\Repeat
		\For{$t=\ell_h$ to $T-\ell+1$}
		\State $[{\bm Z}^h_t, \bm{Z}_{t+1}]\leftarrow \texttt{SeqEnc}({\bm W}^h_t, \bm{W}_{t+1}; \bm{\Theta}_{\text{SE}})$
		\State ${\widehat{\bm Z}}_{t+1}\leftarrow \texttt{Predic}({\bm Z}^h_t;\bm{\Theta}_{\text{PD}})$
		\State $[{\widehat{\bm W}}^h_t, {\widehat{\bm W}}_{t+1}] \leftarrow \texttt{SeqDec}({\bm Z}^h_t, {{\bm Z}}_{t+1};\bm{\Theta}_{\text{SD}})$
		\For{$k=1$ to $K$}
		\State $\bm{\epsilon}_k \sim \mathcal{N}({\bm 0}, {\bm\Sigma})$
		\State ${\widetilde{\bm Z}}_{t+1,k} \leftarrow \texttt{RandPerturb}(\bm{Z}_{t+1}, \widehat{\bm{Z}}_{t+1}, \bm{\epsilon}_k;\bm{\Theta}_{\text{RP}})$
		\State $[*,{\widehat{\bm W}}_{t+1,k}]\leftarrow \texttt{SeqDec}({\bm Z}^h_t, {\widetilde{\bm Z}}_{t+1,k};\bm{\Theta}_{\text{SD}})$
		\EndFor
		\State Update $\bm{\Theta}_{\text{SE}}$, $\bm{\Theta}_{\text{SD}}$, $\bm{\Theta}_{\text{PD}}$ and $\bm{\Theta}_{\text{RP}}$ using Eq. (\ref{equ:Loss})
		\EndFor
		\State $e\leftarrow e + 1$
	\Until{$e=MaxEpoch$}
	\end{algorithmic}
\end{algorithm}


\section{\bf{Experiments on Real-World Datasets}}\label{sec:Experiment}

In this section, we conduct experiments on four public real-world datasets, 
which are extensively used to evaluate anomaly detection algorithms.   
Experiment results show that our proposed LPC-AD algorithm 
outperforms SOTA algorithms 
with respect to both detection accuracy and training time.  

\subsection{Experiment Setting}
\noindent
{\bf Device's Information.} 
We run algorithms on a server, 
which has a CPU (Intel(R) Xeon(R) Platinum 8151 CPU @ 3.40GHz, 96GB memory) 
and a GPU (Quadro GV100, 32GB memory).  

\noindent
{\bf Public Datasets.}  
We use four public datasets which are extensively used in previous works  
\cite{su2019robust,xu2018unsupervised,zhang2019deep,audibert2020usad,tuli2022tranad,li2021multivariate,dai2021sdfvae}.   
Each dataset contains a train set and a testing set.  
Each data point in the testing dataset has a binary label 
indicating whether it is an anomaly (1) or not (0).     
\ref{tab: Dataset Information} summarizes the overall statistics of 
each dataset.  More details of each dataset are described as follows.  

\begin{itemize}
\item {\bf Server Machine Dataset} (SMD\footnote{https://github.com/NetManAIOps/OmniAnomaly/tree/master/ServerMachineDataset}).  
The SMD\cite{su2019robust} time series dataset is collected from a large Internet company.  
Its dimension is 38 and each dimension corresponds to 
one performance metric of a server.  
In total, there are 28 time series and 
each time series corresponds to one server.  
The horizon of each time series is five weeks.   
SMD is divided into two subsets of equal size, 
where the first half is the training set, and the second half with anomaly labels is the testing set. 

\item {\bf Application Server Dataset} (ASD\footnote{https://github.com/zhhlee/InterFusion/tree/main/data/processed}). 
This dataset is also collected from a large Internet company, publiced by \cite{li2021multivariate}.  
Its dimension is 19 and each dimension corresponds to 
one performance metric that characterizes the status of a server (CPU-related metrics and memory-related metrics, etc.).  
In total, there are 12 time series and 
each time series corresponds to one server.  
Each time series records a server's status data in 45 days, with a fixed rate 5 minites.  
The first 30-day-long data  without anomaly labels are used for training, 
and the last 15-day-long data with anomaly label are used for anomaly testing.

\item{\bf Secure Water Treatment}(SWaT\footnote{https://github.com/JulienAu/Anomaly\_Detection\_Tuto}).  
This dataset is from a real-world industrial water treatment plant producing filtered water \cite{mathur2016swat}.  
The SWaT dataset is a scaled-down version of the original data.  
Its dimension is 51, and each dimension corresponds to one performance metric of the 
industrial water treatment plant.  
The time horizon of this time series is 11 days.  
The dataset in the first seven days is collected under normal operations, and that in the last four days is collected with attacks.  
 
\item{\bf Water Distribution Dataset}(WADI\footnote{https://itrust.sutd.edu.sg/itrust-labs\_datasets/dataset\_info/\#wadi}). 
This dataset is collected from the WADI testbed, 
and it is an extension of the SWaT\cite{ahmed2017wadi} dataset. 
Its dimension is 123, and each dimension corresponds to one performance metric of the 
industrial water treatment plant.  
It only contains one time series corresponding to one server.  
The time horizon of this time series is 16 days.  
The dataset in the first 14 days is collected under normal operations, 
and that in the last two days is collected with attacks.   
\end{itemize}

\begin{table}[h]
  \caption{Statistics of four public datasets.}
  \label{tab: Dataset Information}
  \center
  \begin{tabular}{cccccc}
    \toprule
    	Dataset & \specialcell{\# time\\ series} & \# metrics & \# train & \# test & \# anomaly(\%)\\
    \midrule
     SMD & 28 & 38 & 708405 & 708420 & 4.16\\
     ASD  & 12 & 19 & 102331 & 51840 & 4.61\\
     WADI & 1 & 123 & 784571 & 172801 & 5.77\\
	SWaT & 1 & 51 & 475200 & 449919 & 12.13\\
    \bottomrule
  \end{tabular}
\end{table}

\noindent
{\bf Comparison baselines.}  
To show the merit of our proposed algorithms, 
we consider the following SOTA baselines.    
\begin{itemize}[noitemsep,topsep=0pt,leftmargin=*]
\item 
{\bf OmniAnomaly}\footnote{OmniAnomaly open source link: https://github.com/NetManAIOps/OmniAnomaly}\cite{su2019robust}.  
This method is built from a variational autoencoder.  
Its sophisticated model 
to capture temporal dependence of time series
 enhances the detection accuracy,
 but leads to a slow training speed.  
  
\item 
{\bf InterFusion}\footnote{InterFusion open source link: https://github.com/zhhlee/InterFusion}\cite{li2021multivariate}.  
This method is an improved variant of OmniAnomaly.  
The improvement is on the detection accuracy of OmniAnomaly.
The model is still sophisticated and the training speed is slow.  
 
\item 
{\bf USAD}\footnote{USAD open source link: https://github.com/manigalati/usad}\cite{audibert2020usad}.  
This method achieves a fast training speed 
by trading-off the detection accuracy.  
It is built on autoencoder and uses adversarial training.   

\item 
{\bf TranAD}\footnote{TranAD open source link: https://github.com/imperial-qore/TranAD}\cite{tuli2022tranad}.  
Similar to USAD, this method focuses on a fast training speed by trading-off detection accuracy.  
It is built on the transformer architecture.   
\end{itemize}
We do not compare with a number of other notable baselines such as 
\cite{Deng2021,li2019mad,Zhao2020,zong2018deep}, 
due to that they are shown inferior to TranAD and 
we omit then for brevity and simplicity of presentation.  

To reveal a fundamental understanding of our proposed LPC-AD framework, 
we consider the following instances of it, 
where different instances have differet predictors.  
\begin{itemize}[noitemsep,topsep=0pt,leftmargin=*]
\item 
{\bf LPC-AD-SA} 
sets the predictor function $\texttt{Predic}(\cdot;\bm{\Theta}_{\text{PD}})$ 
to be an attention enabled Seq2Seq shown in Figure \ref{fig:seq2seq-att}.

\item 
{\bf LPC-AD-S} 
sets the predictor function $\texttt{Predic}(\cdot;\bm{\Theta}_{\text{PD}})$ 
to be a LSTM enabled Seq2Seq shown in Figure \ref{fig:seq2seq}.

\item 
{\bf LPC-AD-L}
 sets the predictor function $\texttt{Predic}(\cdot;\bm{\Theta}_{\text{PD}})$ 
as a linear function shown in Equation (\ref{Predictor:LinearTransfor}).
\end{itemize}

\noindent
{\bf Evaluation Metrics}.   
Following previous works \cite{su2019robust,xu2018unsupervised,zhang2019deep,audibert2020usad,tuli2022tranad,li2021multivariate,dai2021sdfvae}, 
we use precision, recall, area under the receiver operating characteristic curve (AUROC), and F1 score 
to evaluate the detection accuracy of all algorithms.  
We use these metrics that are designed for binary classification because
anomaly detection has a binary output, i.e., anomaly (1) or not (0).  

In real-world applications, 
 anomalous data points often occur consecutively as an anomalous segment.   
The purpose of anomaly detection is to notify system operators about the possible issues.  
Thus, it is acceptable that an anomaly detection algorithm can 
trigger the anomaly alarm at any subset of a segment of anomalies, 
instead of correctly classifying each anomalous data points.  
For this practical consideration, 
we use a point-adjustment strategy proposed by \cite{xu2018unsupervised} to calculate the performance metrics.  
This point-adjustment strategy was widely recognized and applied in previous works \cite{su2019robust,zhang2019deep,audibert2020usad,tuli2022tranad,li2021multivariate,dai2021sdfvae}.   
According to this strategy, if at least one data point of an anomalous segment is 
classified as an anomaly, all the data points of this anomalous segment is 
considered to be correctly classified as an anomaly.    

There are randomness in the training process of these algorithms. 
The randomness comes from random sampling and 
random initialization.    
Faced with such randomness, we repeat the training and testing of each algorithm for 
$D \in \mathbb{N}_+$ times, 
and take the average results.
In some datasets, the number of time series $N$ is more than one.  
For example, SMD contains 28 time series.  
To simplify the presentation, we consider the average of the performance metric 
across all time series in a dataset.   
Formally, let $P_{i,j}, R_{i,j}, F_{1,i,j}, AUROC_{i,j}$ denote the 
precision, recall, $F_1$ score and receiver operating characteristic curve 
of the an algorithm on the $i$-th time series of a dataset 
in the $j$-th repeated training and testing.  

\begin{align}
& 
P = \frac{1}{ND}\sum\nolimits^{N}_{i=1}\sum\nolimits^{D}_{j=1} P_{i,j}, 
\quad
R = \frac{1}{ND}\sum\nolimits^{N}_{i=1}\sum\nolimits^{D}_{j=1} R_{i,j}, \\
& 
F_1=\frac{1}{ND}F_{1,i,j}, 
\quad
F^*_1=2\cdot\frac{P \times R}{P + R},\\
& 
AUROC =  \frac{1}{ND}\sum\nolimits^{N}_{i=1}\sum\nolimits^{D}_{j=1} AUROC_{i,j}.
\end{align}
Notably, $F_1$ is called the micro $F_1$ score 
and $F^\ast_1$ is called the macro $F_1$ score.  
In our experiments, we conduct $D=8$ rounds of repeated 
training and testing, because some baseline algorithms is time-consuming.

\noindent
{\bf Hyperparameter setting \& implementation details.}  
For the baseline algorithms (OmniAnomaly, InterFusion, USAD and TranAD),  
we set their hyperparameters such as window size, embedding size, etc., 
according to 
their paper or 
their open-source code
 (when it is not stated in the paper).  
For three instances of LPC-AD, i.e., LPC-AD-SA, LPC-AD-S and LPC-L, 
we use the following default hyper parameters: 
historical window size $\ell_h = 10$, 
future window size $\ell = 2$, 
hidden layer dimension of LSTM = $M / 2$, 
embedding dimension $N = 8 (16 \text{ for WADI dataset})$, 
covariance matrix $\bm{\Sigma} = \bm{I}$, 
learning rate $=0.001$, 
maximum training epoch $MaxEpoch=40 (25 \text{ for WADI dataset})$, 
training batch size$=64$.  
We also vary the hyperparameters to study their 
impact on the detection accuracy and training time. 
By default, we use 100\% training dataset for all the algorithms.
We implemented instances of LPC-AD with 
PyTorch-1.9.0 library, 
trained with Adam optimizer.

All algorithms in consideration detect anomalies based on the alert threshold $\lambda$.  
Similar with previous works \cite{su2019robust,zhang2019deep,audibert2020usad,tuli2022tranad,li2021multivariate,dai2021sdfvae}, 
we use exhaustive search to select the optimal alert threshold.  
In the search, the objective is to achieve the highest $F_1$ metric.    
For instances of LPC-AD, 
we exhaustively search in [0,1] with a step size of 0.0001.
For other baseline algorithms, we use the exhaustive search method stated in their paper.  

\subsection{Comparison with SOTA Baselines}
 
We first consider the setting where all training data are used for training.  
We will study the impact of training data size later.  
We start our evaluation by answering the following question: 

\noindent
\textit{\bf Q1: Can LPC-AD improve the detection accuracy and training time compared to the SOTA baselines?}

Consider two baselines with sophisticated models with SOTA detection accuracy.  
Compared with OmniAnomaly, our LPC-AD-SA improves 
its $F_1$ and $F^\ast_1$ by at least 11.2\% over the ASD, WADI and SWaT datasets 
and by 2.2\% over the SMD dataset.  
It is worth noting that the on the WADI dataset, 
our LPC-AD-SA improves the $F_1$ and $F^\ast_1$ of OmniAnomaly by around 57.4\%.  
This drastic improvement is due to the fact that the WADI dataset is highly non-smooth, 
over which the sophisticated model of OmniAnomaly overfits the data. 
Compared with InterFusion, which is an improved variant of OmniAnomaly, 
our LPC-AD-SA improves 
its $F_1$ and $F^\ast_1$ by at most 18.9\% over the WADI and SWaT datasets 
and by at least 2.1\% over the SMD and ASD dataset.  
Namely, LPC-AD-SA significantly outperform the OmniAnomaly and InterFusion  
in terms of $F_1$ and $F^\ast_1$
For the AUROC accuracy measure,  LPC-AD-SA also 
significantly outperform the OmniAnomaly and InterFusion.  

Furthermore, our LPC-AD-SA can significantly outperform these baselines 
in terms of training speed as well.  
In particular, Table \ref{tab: Training Time} shows that the per epoch training time 
of our LPC-AD-SA is less than 10\% of that of both OmniAnomaly and InterFusion over four datasets.   
In other words, the training time of OmniAnomaly and InterFusion are 
roughly an order of magnitude larger than our LPC-AD-SA.   

\noindent{\bf Answer 1.1:} 
\emph{
	LPC-AD-SA improves the detection accuracy of both OmniAnomaly and InterFusion 
significantly, and it reduces the training time of both OmniAnomaly and InterFusion drastically.  
}

\begin{table}[htb]
\center
  \caption{Comparison of detection accuracy with respect to  
  precision($P$), recall($R$) , AUROC, $F_1$ and $F^*_1$ score.}
  \label{tab:Overview result}

  \begin{tabular}{lccccc}
    \toprule
     \multirow{2}{*}{Method} & \multicolumn{5}{c}{SMD}\\
	\cline{2-6}
	& P & R & AUROC & $F_1$ & $F^*_1$\\ \hline
	OmniAnomaly & {\bf 0.9599}	& 0.9091	& 0.9539	& 0.9275	& 0.9338\\ 
	InterFusion & 0.9262 & 0.9394 & 0.9939 & 0.9287 & 0.9328\\ 
	USAD & 0.8968 & 0.9054 & 0.9499 & 0.8874 & 0.9011\\ 
	TranAD & 0.9475 & 0.9465 & 0.9954 & 0.9414 & 0.947\\ 
	{\bf LPC-AD-SA} & 0.9401 & \bf 0.967 & \bf 0.9973 & \bf 0.9483 & \bf 0.9533\\ 
	\hline    
     \multirow{2}{*}{Method} & \multicolumn{5}{c}{ASD}\\
	\cline{2-6}
	& P & R & AUROC & $F_1$ & $F^*_1$\\ \hline
	OmniAnomaly & 0.8469 & 0.8593 & 0.9801 & 0.8348 & 0.8531\\ 
	InterFusion & 0.9014 & \bf 0.9734 & 0.997 & 0.9342 & 0.936\\ 
	USAD & \bf 0.9474 & 0.8742 & 0.9906 & 0.8989 & 0.9093\\ 
	TranAD & 0.8858 & 0.9027 & 0.9846 & 0.8813 & 0.8942\\ 
	{\bf LPC-AD-SA} & 0.9201 & 0.9539 & \bf 0.998 & \bf 0.935 & \bf 0.9367\\ 
	\hline    
     \multirow{2}{*}{Method} & \multicolumn{5}{c}{WADI}\\
	\cline{2-6}
	& P & R & AUROC & $F_1$ & $F^*_1$\\ \hline
	OmniAnomaly 		& 0.6746 		& 0.4587 		& 0.7225 		& 0.537 		& 0.5461\\ 
	InterFusion 		& 0.7977 		& 0.6589 		& 0.9303 		& 0.7114 		& 0.7217\\ 
	USAD 			& 0.9293 		& 0.535 		& 0.8966 		& 0.6649 		& 0.6791\\ 
	TranAD 			& \bf 0.984 	& 0.2405 		& 0.6864 		& 0.3865 		& 0.3865\\ 
	{\bf LPC-AD-SA} 		& 0.9369 		& \bf 0.7745 	& \bf 0.9755 	& \bf 0.8455 	& \bf 0.8481\\ 
	\hline    
     \multirow{2}{*}{Method} & \multicolumn{5}{c}{SWaT}\\
	\cline{2-6}
	& P & R & AUROC & $F_1$ & $F^*_1$\\ \hline
	OmniAnomaly 	& 0.7241 		& 0.756 	& 0.8519 	& 0.7256 	& 0.7397\\ 
	InterFusion 	& 0.8495		&0.8292	 	& 0.9363	& 0.814		&0.8392\\ 
	USAD 			& 0.9581 		& 0.8625 	& 0.9686 	& 0.9074 	& 0.9078\\ 
	TranAD 			& \bf 0.9789 	& 0.6963 	& 0.9338 	& 0.8138 	& 0.8138\\ 
	{\bf LPC-AD-SA} 	& 0.961 		& \bf 0.9376 		& \bf 0.9926 & \bf 0.9489 & \bf 0.9492\\ 
    \bottomrule
  \end{tabular}

\end{table}

\begin{table}[htb]
\center
  \caption{Comparison of per epoch training time (in seconds).}
  \label{tab: Training Time}
  \begin{tabular}{lcccc}
    \toprule
    	 & SMD & ASD & WADI & SWaT \\
    \midrule
     OmniAnomaly & 2945.3 & 293.4 & 3937.1 & 2956.4 \\
     InterFusion & 2891.6 & 397.1 & 2432.8 & 2076.6\\
     TranAD & 304.8 & 27.3	& 275.2 & 168.5\\
	 USAD & 229.7 & 31.2 & 252.3 & 161.2 \\
	 {\bf LPC-AD-SA} & \bf{188.1} & \bf{26.4} & \bf{209.7} & \bf{118.2}\\
    \bottomrule
  \end{tabular}
  \vspace{-0.08in}
\end{table}

Consider two baselines with simple models and SOTA training speed, i.e., USAD and TranAD.  
Table \ref{tab: Training Time} shows that LPC-AD-SA has a shorter 
per epoch training time than both TranAD and USAD.  
More specifically, compared with TranAD, our LPC-AD-SA reduces its per epoch running 
time by at most 38.2\% and by at least 24\% over the SMD, WADI and SWaT datasets.  
Over the ASD dataset, LPC-AD-SA reduces the per epoch training time 
of TranAD by 3.7\%.  
One reason to have this small reduction is that the number of dimension of the ASD dataset is small, i.e., 19, 
thus TranAD is already fast enough, 
leaving a small room for further improvement.   
Compared with USAD, our LPC-AD-SA reduces its per epoch running 
time by 16\%-27\% over four datasets.  
One reason to have this large reduction in training time over all datasets is that 
the USAD is not fast enough even when the dimension of a dataset is small, 
leaving a large room for further improvement.   
These results show that our LPC-AD-SA significantly reduces the running time of TranAD and USAD.  
Moreover, as shown in Table \ref{tab:Overview result}, 
our LPC-AD-SA can also significantly improve the detection accuracy 
of TranAd and USAD.  
In particular, consider the WADI dataset, 
our LPC-AD-SA improves  the $F_1$ and $F^\ast_1$ of both TranAD and USAD by 
24.9\%-119.42\%.
On the WADI dataset, LPC-AD-SA improves the AUROC of USAD and TranAD by 8.7\% and 42.1\%, respectively.   
One reason to have this drastic improvement is that the WADI is highly non-smooth, 
over which simple models of TranAD and USAD may under-fit the data, 
leading to low detection accuracy and leaving a large room for further improvement.  
For the other three datasets, the time series are smoother than WADI, 
over which LPC-AD-SA has a smaller improvement (i.e., several percent) of $F_1$, $F^\ast_1$ and AUROC over 
the TranAD and USAD than the WADI dataset.  
The reason is that over these datasets, the accuracy of  TranAD and USAD is already not low, 
leaving limited room for further improvement.  

\noindent{\bf Answer 1.2:} 
\emph{
	LPC-AD-SA improves the detection accuracy of both USAD and TranAD 
significantly, especially when the time series data is highly non-smooth. It also reduces the training time.  
}

\noindent
{\bf Q2: Is the improvement robust to training dataset size?}  

In order to further validate the superior performance of our LPC-AD-SA 
over SOTA baselines, we study the impact of training dataset size on the 
detection accuracy and training time of the algorithms. 
We use the following method to select a subset of the training dataset.  
Note that all the algorithms in comparison slice the training time series dataset into sliding windows 
and then treat each two consecutive window pair as one training data item.  
When we say a fraction of the training data, such as 50\%, 
we mean randomly selecting 50\% of the training data items (i.e. sliding window pairs).   
This randomness in the selection of training data 
leads to uncertainty in the output of the algorithm.  
To eliminate this uncertainty, we repeat each algorithm 
eight times and take the average as the final output.  
To simplify the presentation, here, we only consider the SMD dataset.  
The reason is that, as shown in Table \ref{tab:Overview result}, 
the accuracy improvement of LPC-AD-SA over four SOTA baselines 
is roughly the smallest among these four datasets.  
If LPC-AD-SA can still outperform four SOTA baselines in SMD 
under different training dataset sizes, 
then one can expect higher improvement over the other three datasets.  
Another reason is that the SMD datasets has a relatively larger number of time series 
and has rich temporal patterns in the datasets.  

\begin{figure}[htb]
	\centering
	\subfigure[$F_1$ score]{
		\centering
		\includegraphics[width=0.225\textwidth]{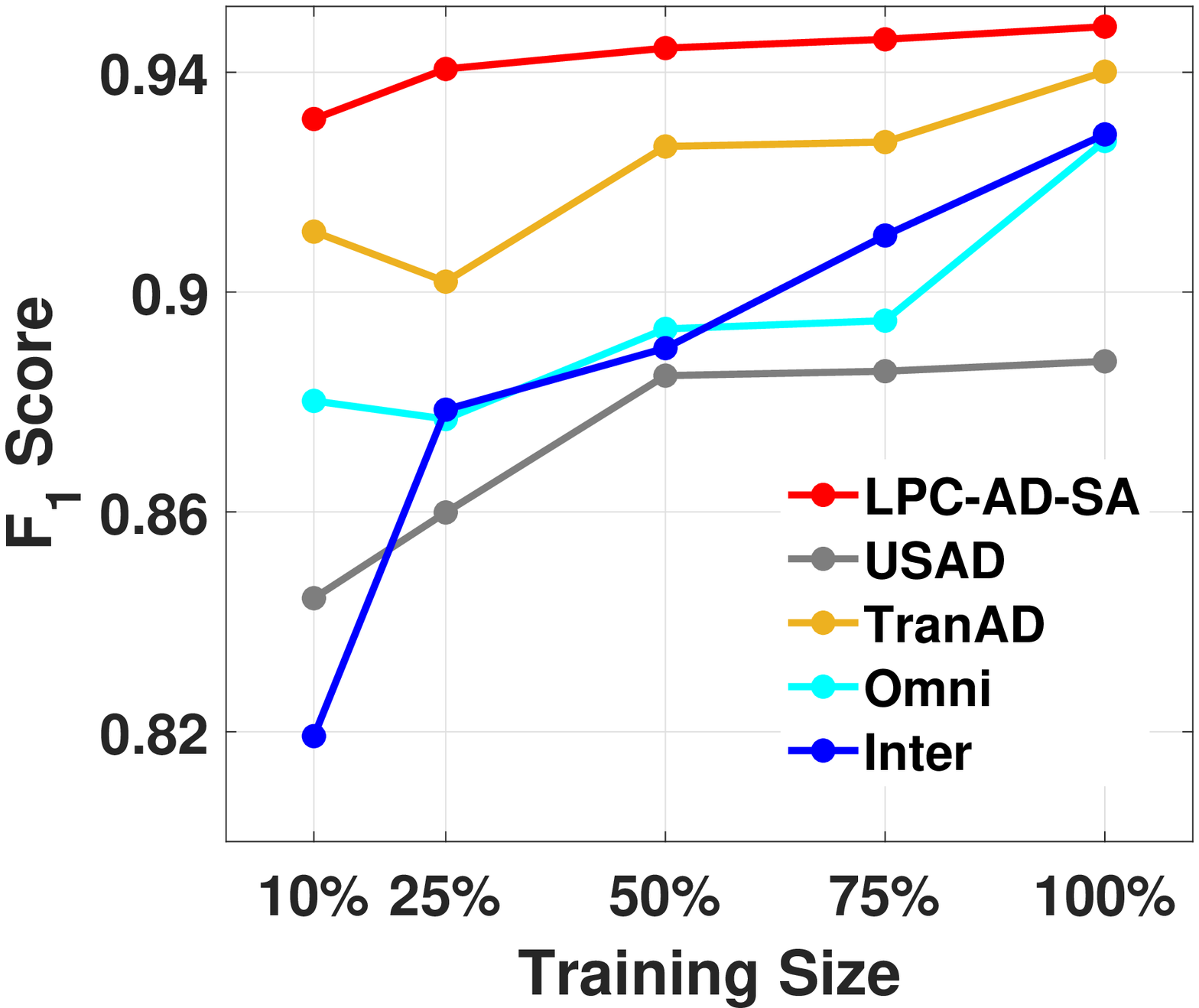}
		\label{fig: TrainSize SMD-OF1}
	}
	\subfigure[$F^*_1$ score]{
		\centering
		\includegraphics[width=0.225\textwidth]{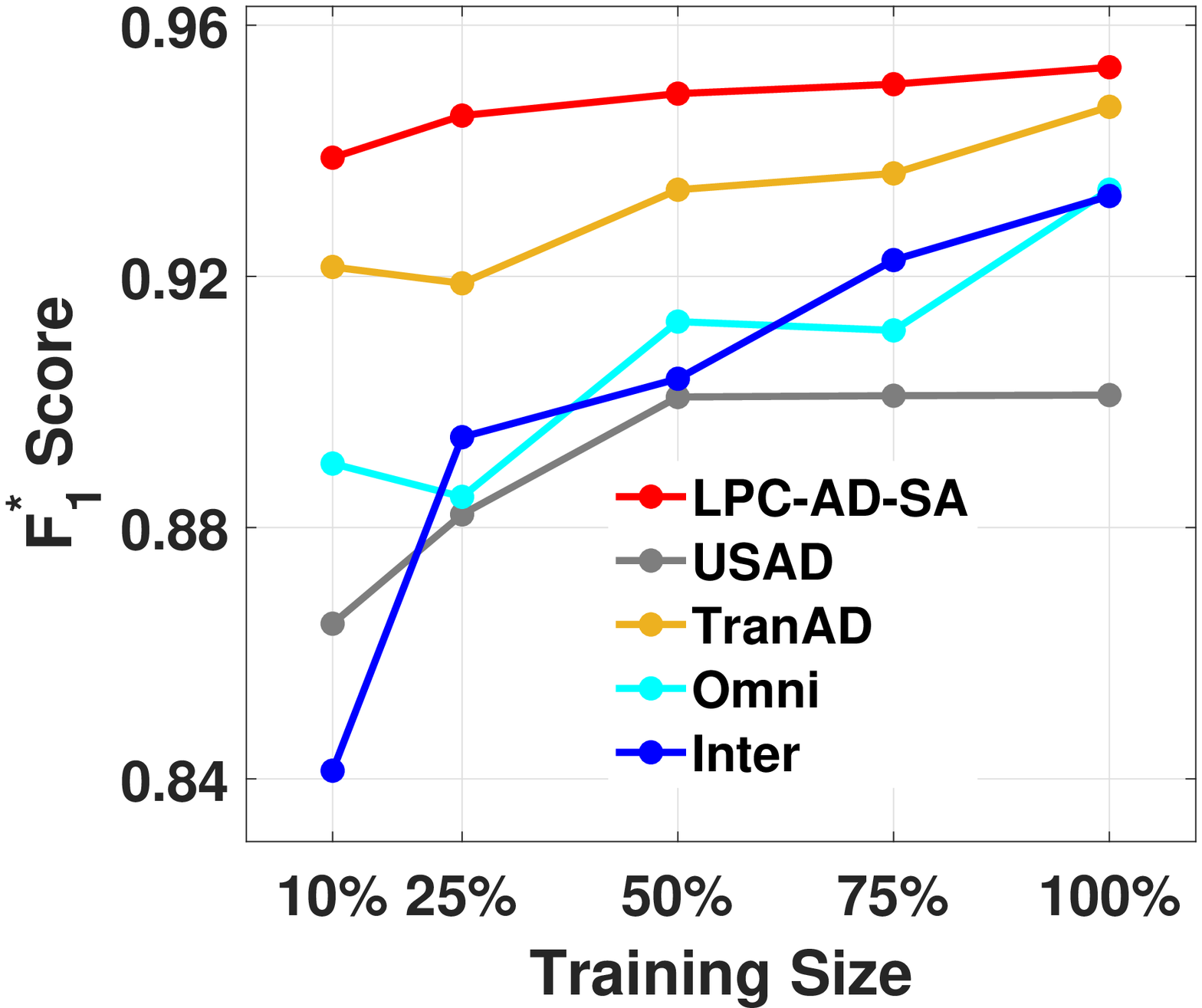}
		\label{fig: TrainSize SMD-F1}
	}
	\subfigure[AUROC score]{
		\centering
		\includegraphics[width=0.225\textwidth]{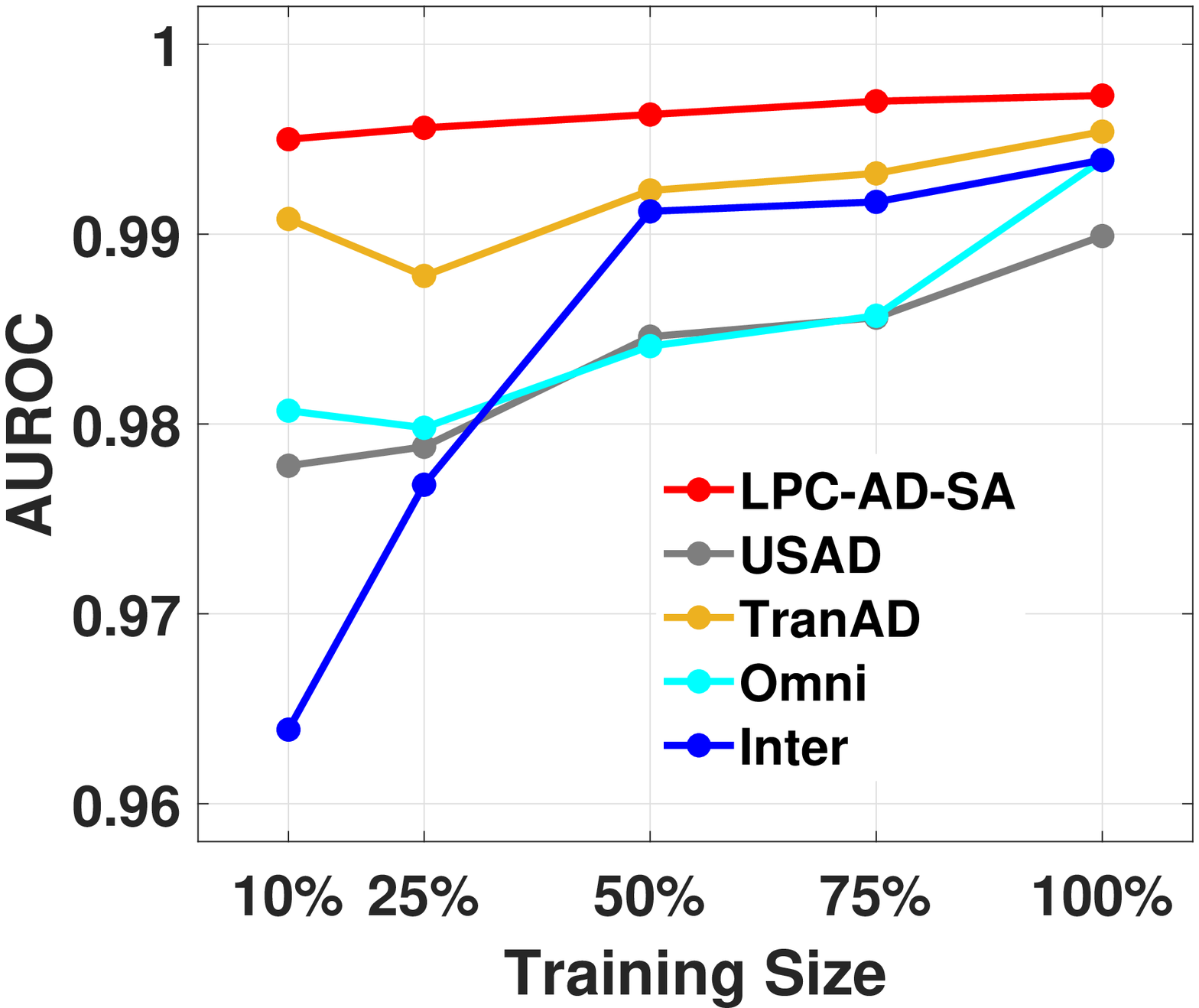}
		\label{fig: TrainSize SMD-AUROC}
	}
	\subfigure[Training time]{
		\centering
		\includegraphics[width=0.225\textwidth]{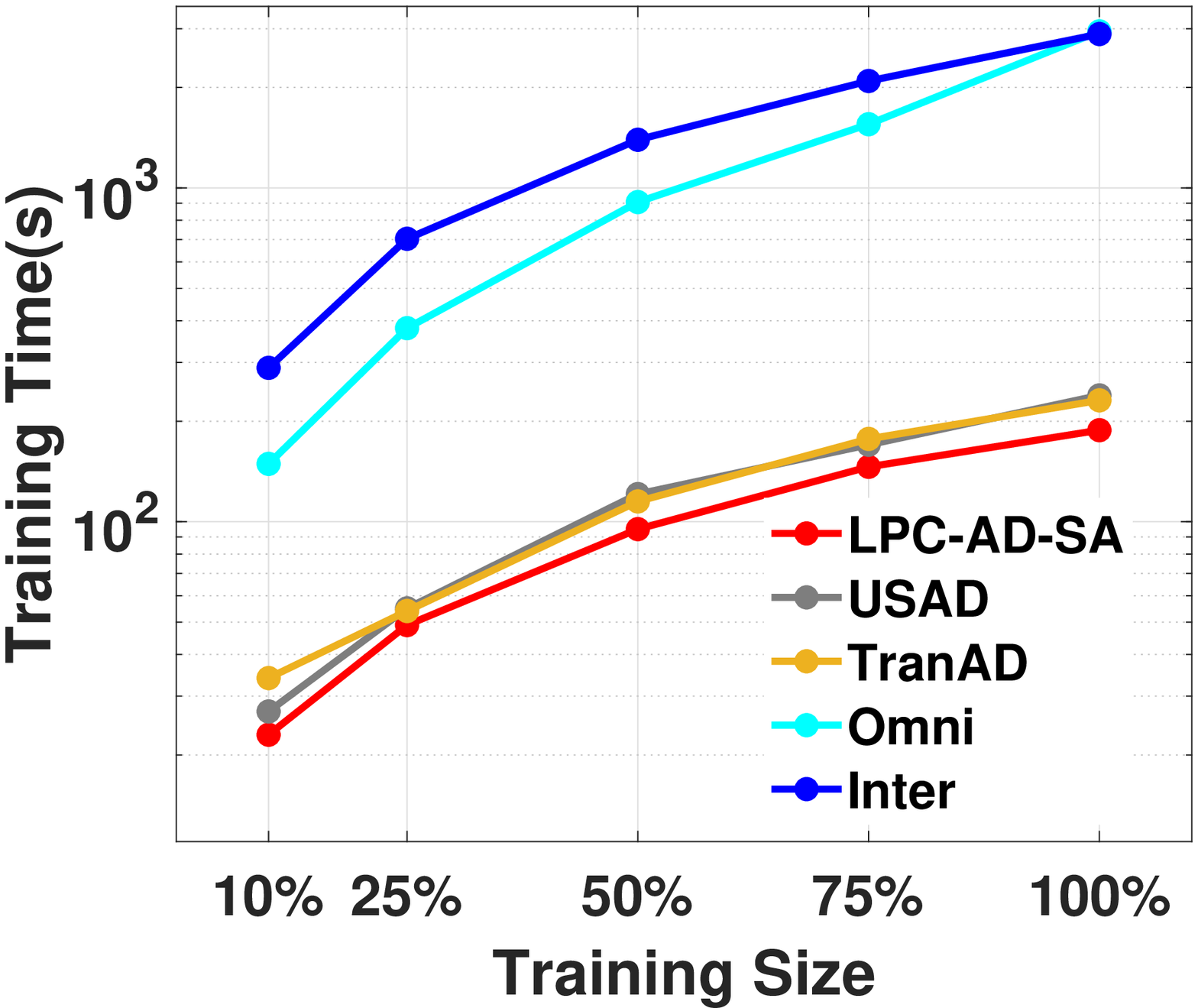}
		\label{fig: TrainSize SMD-TrainingTime}
	}
	\caption{Impact of training dataset size on the anomaly detection accuracy and training time.}  
	\label{fig:SMD Training Size} 
\end{figure}

Figure \ref{fig: TrainSize SMD-OF1} shows that as the fraction of the training dataset increases 
from 10\% to 100\%, the $F_1$ score of LPC-AD-SA and four SOTA baselines increases.  
Furthermore, the $F_1$ curve of LPC-AD-SA  lies in the top.  
This means that LPC-AD-SA has higher $F_1$ than four SOTA baselines 
under different training dataset sizes.  
The $F_1$ curve of LPC-AD-SA is quite flat.  
This shows that the training accuracy of LPC-AD-SA is robust to training dataset size.  
This robustness validates the power of LPC-AD-SA in learning 
the normal temporal dependence.  
Figure \ref{fig: TrainSize SMD-F1} and \ref{fig: TrainSize SMD-AUROC} 
shows similar finding 
with respect to the accuracy measure $F^\ast_1$ and AUROC.  
Figure \ref{fig: TrainSize SMD-TrainingTime} shows that 
as the fraction of the training dataset increases 
from 10\% to 100\%, the training time of LPC-AD-SA and four SOTA baselines increases 
in a nearly linear rate.  
The training time curve of LPC-AD-SA lies at the bottom.  
This means that LPC-AD-SA has the fastest training speed than four SOTA baselines 
under different training dataset sizes.  
The improvement ratio of train time by the LPC-AD-SA or SOTA baselines 
across different training dataset sizes varies slightly.  
In summary, the accuracy of LPC-AD-SA is robust to training dataset size.  
LPC-AD-SA outperforms four SOTA baselines significantly 
in terms of both detection accuracy and training time under 
different selections of training dataset size.  

\noindent{\bf Answer 2:}
\emph{For different training dataset size, LPC-AD-SA robustly improves the detection accuracy and training time over the baselines. }


\subsection{Parameter Sensitivity Analysis}

To reveal fundamental understandings on our proposed LPC-AD algorithm, 
we now study the anomaly detection accuracy and training time of three instances of LPC-AD 
under different selections of hyperparameters.  
Similar with previous works \cite{su2019robust,zhang2019deep,audibert2020usad,tuli2022tranad,li2021multivariate,dai2021sdfvae}, 
we focus on one dataset, i.e., SMD, for the brevity of presentation.   
One reason for selecting SMD is that the SMD dataset has a relatively larger number of time series 
and has rich temporal patterns in the datasets.  

\noindent
{\bf Q3: Is LPC-AD robust under different parameter settings?}  

\begin{figure}[htb]
	\centering
	\subfigure[$F_1$ score]{
		\centering
		\includegraphics[width=0.225\textwidth]{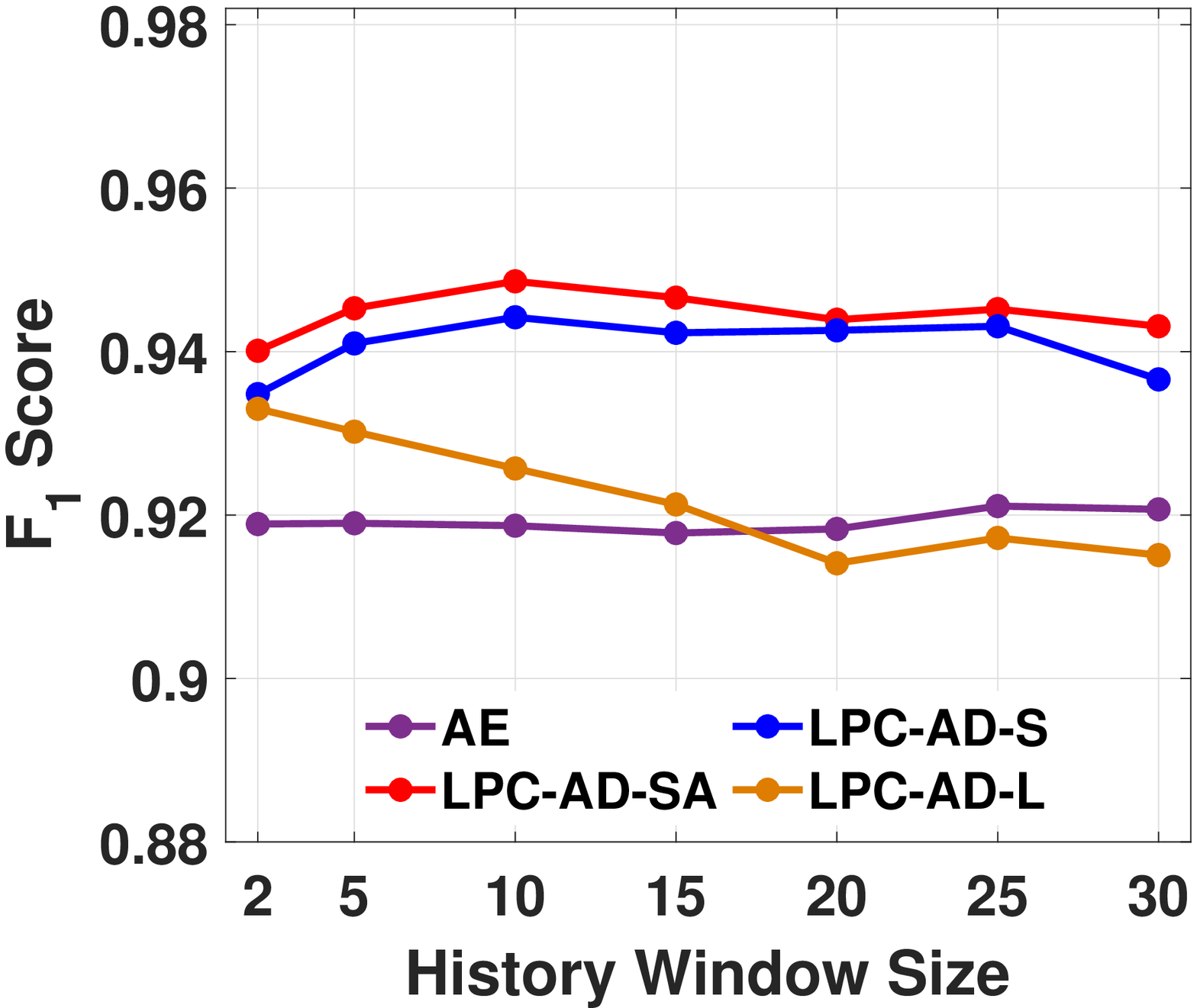}
		\label{fig:WindowSize-OF1}
	}
	\subfigure[$F^*_1$ score]{
		\centering
		\includegraphics[width=0.225\textwidth]{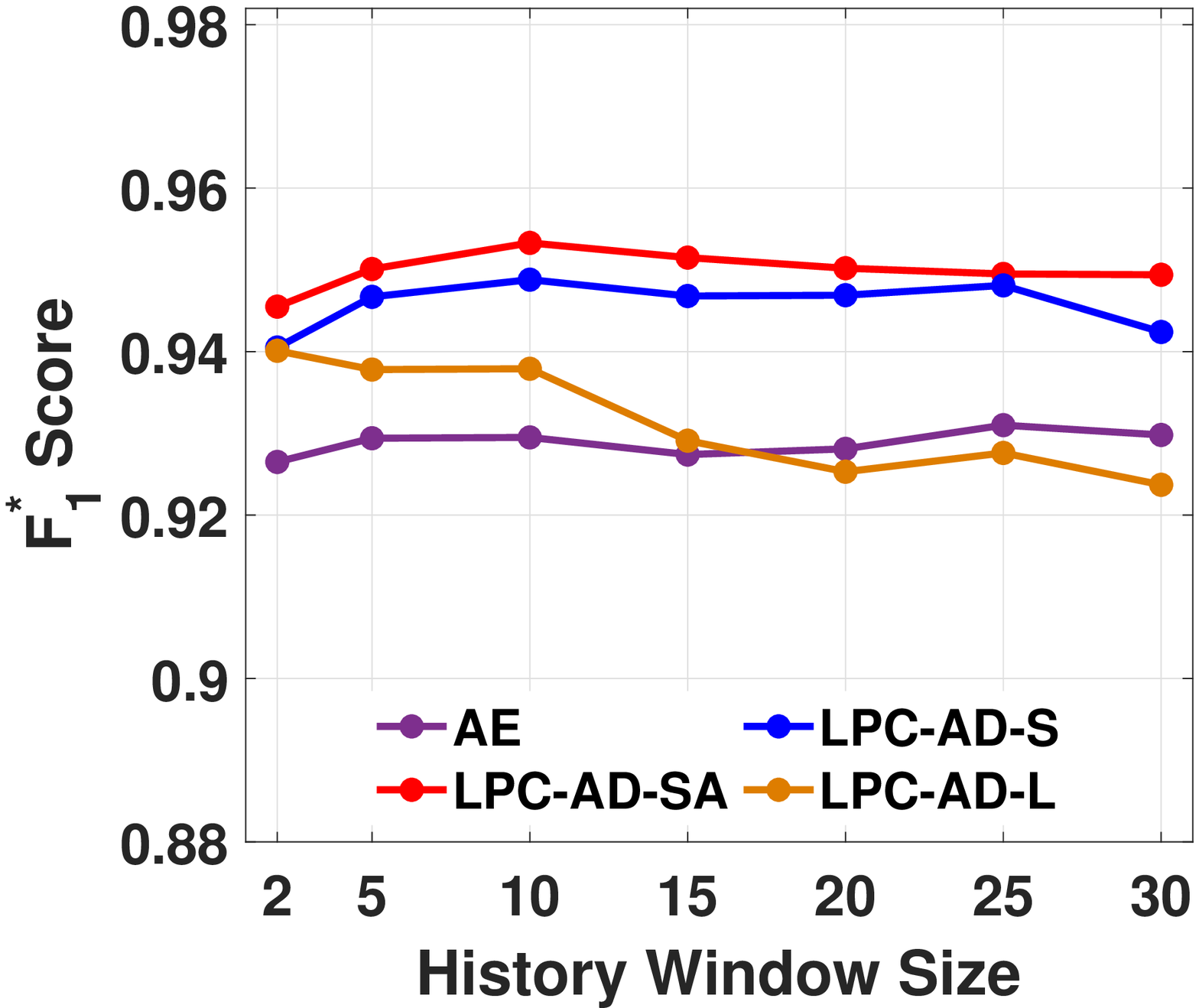}
		\label{fig:WindowSize-F1}
	}
	\subfigure[AUROC score]{
		\centering
		\includegraphics[width=0.225\textwidth]{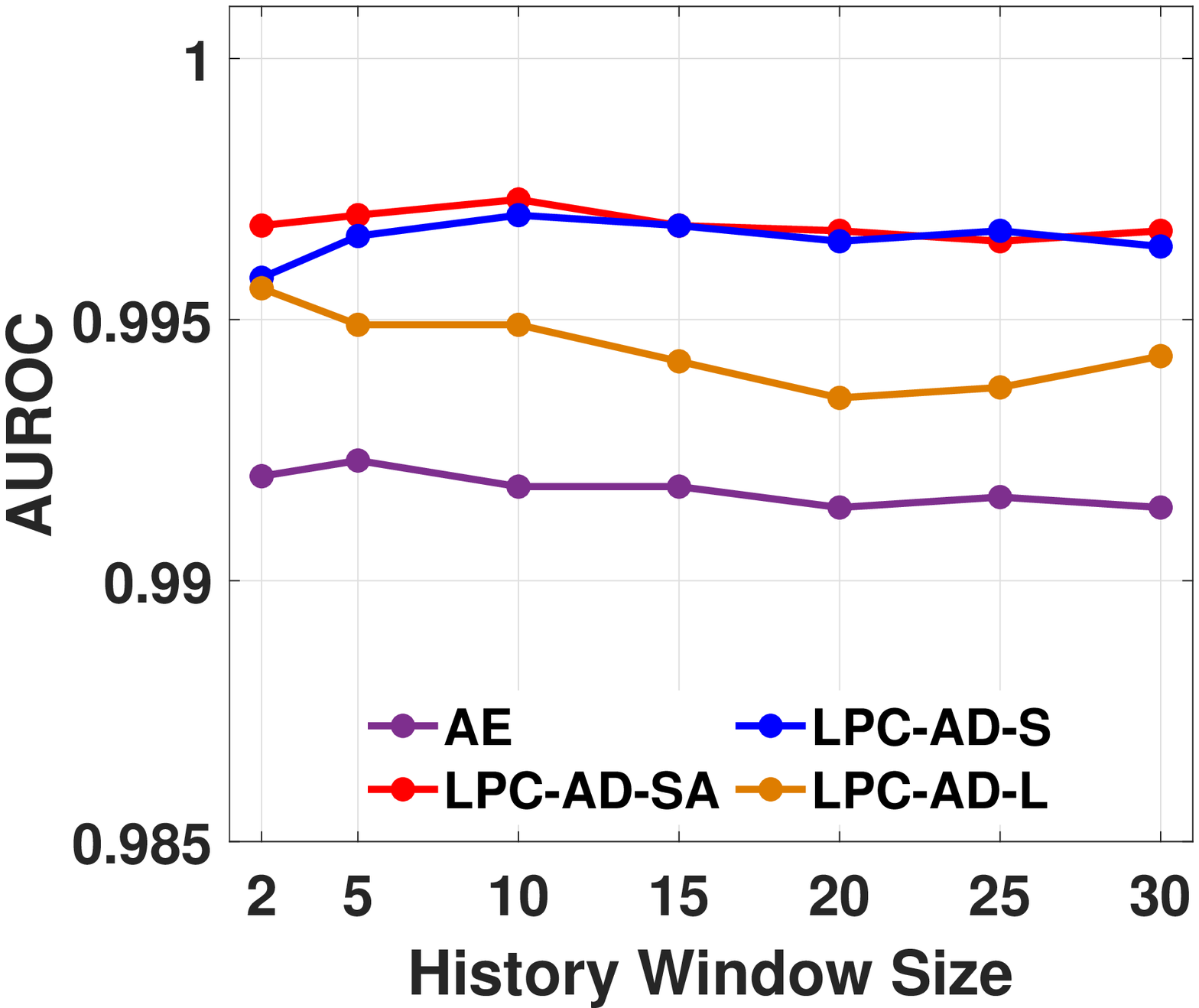}
		\label{fig:WindowSize-AUROC}
	}
	\subfigure[Training time]{
		\centering
		\includegraphics[width=0.225\textwidth]{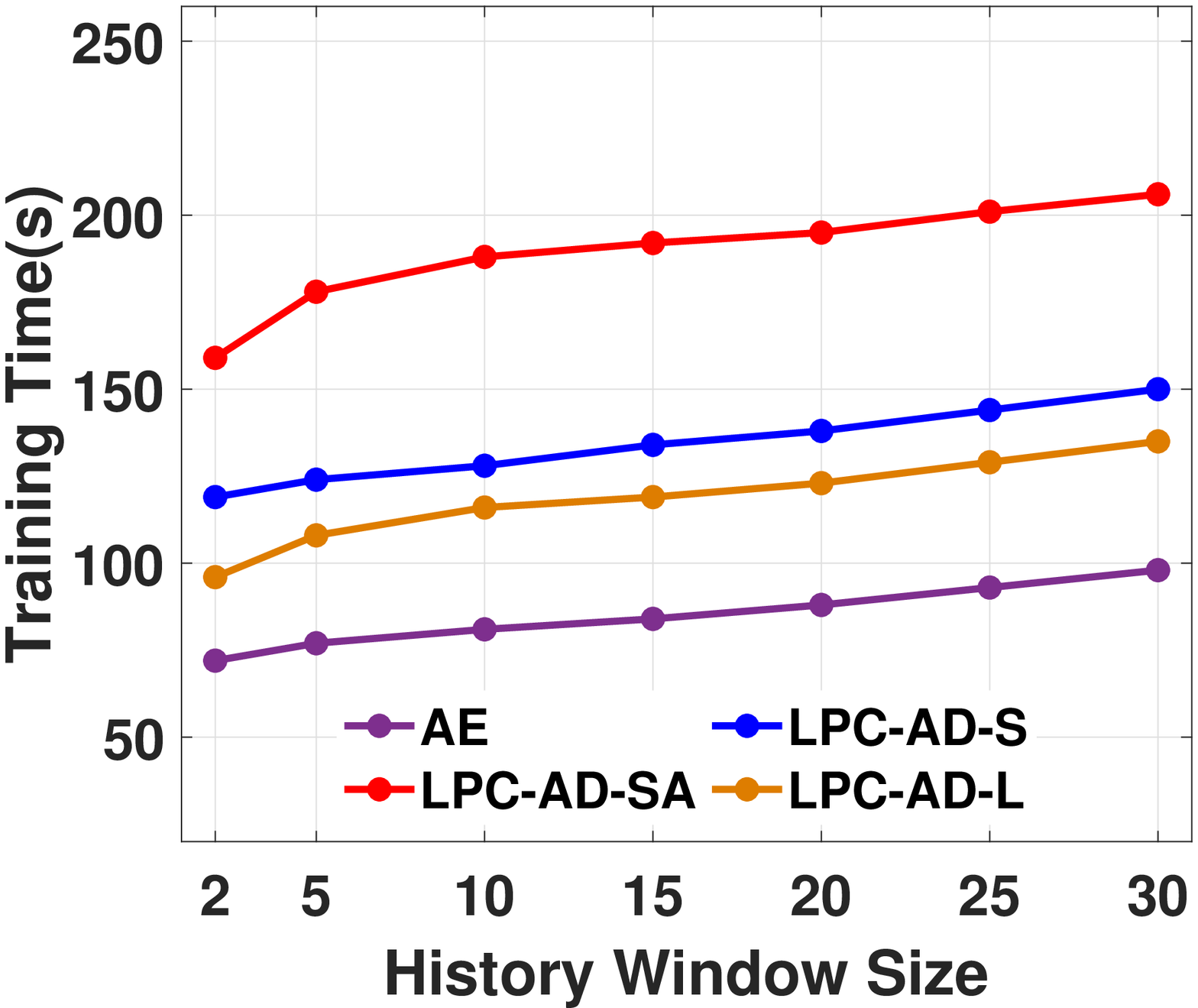}
		\label{fig:WindowSize-TrainingTime}
	}
	\caption{Impact of historical window size $\ell_h$ on the detection accuracy and training time.} 
	\label{fig:SMD First window Size}
	\vspace{-0.06in}
\end{figure}

\noindent
{\bf $\bullet$ Impact of history window size.}  
Figure \ref{fig:SMD First window Size} shows the impact of 
historical window size $\ell_h$ on the detection accuracy and training time 
on the AutoEncoder baseline (AE) and three instances of LPC-AD.  
Figure \ref{fig:WindowSize-OF1} shows that 
the $F_1$ curves of LPC-AD-SA, LPC-AD-S and AE are roughly flat as the historical window size $\ell_h$ 
increases from 2 to 3.  
The same findings can be observed on both the $F^\ast_1$ 
and AUROC metric, 
as shown in Figure \ref{fig:WindowSize-F1} and \ref{fig:WindowSize-AUROC}.  
Namely, the detection accuracy of LPC-AD-SA, LPC-AD-S and AE 
is not sensitive to the historical window size.   
Moreover, from Figure \ref{fig:WindowSize-OF1}-\ref{fig:WindowSize-AUROC},
we observe that the detection accuracy of the linear variant LPC-AD-L 
has a larger variation 
as the historical window size $\ell_h$ changes.  
Note that the detection accuracy of each algorithm is not 
monotone in the historical window size.  
This phenomena that the detection accuracy is not monotone in 
window size is also observed in the experiments of previous works \cite{audibert2020usad,li2021multivariate,tuli2022tranad}.
One reason of this non-monotonicity is that increasing the historical window size 
increases the parameters of the model and it does not increase 
the volume of training data.  
Figure \ref{fig:WindowSize-TrainingTime} shows that the training time 
of LPC-AD-SA, LPC-AD-S, LPC-AD-L and AE 
increases slightly in the historical window size $\ell_h$ 
nearly at a linear rate.  
This implies that these algorithms scale well with respect 
to the historical window size $\ell_h$.   
Furthermore, as shown in Figure~\ref{fig:WindowSize-AUROC}, the gap between $F_1$ (or AUROC) curves of 
 LPC-AD-SA and  LPC-AD-L are quite large.  
This means that capturing the nonlinear dependence 
in the time series data can improve the anomaly detection accuracy significantly.  
Figure \ref{fig:WindowSize-TrainingTime} shows that 
this improvement in the detection accuracy is achieved 
at the cost of slowing down the training speed.

\begin{figure}[htb]
	\centering
	\subfigure[$F_1$ score]{
		\centering
		\includegraphics[width=0.225\textwidth]{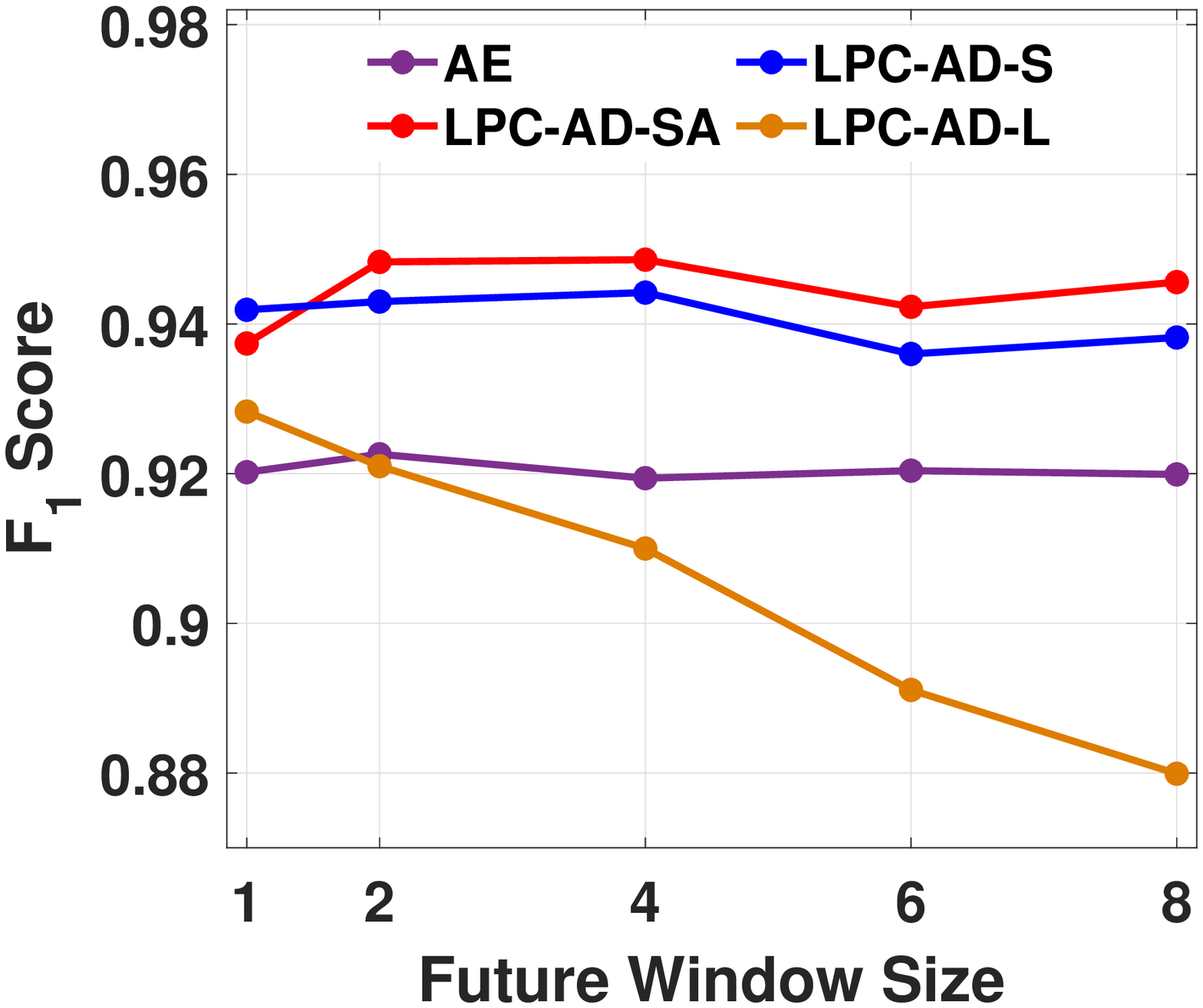}
		\label{fig:WindowSize2-OF1}
	}
	\subfigure[$F^*_1$ score]{
		\centering
		\includegraphics[width=0.225\textwidth]{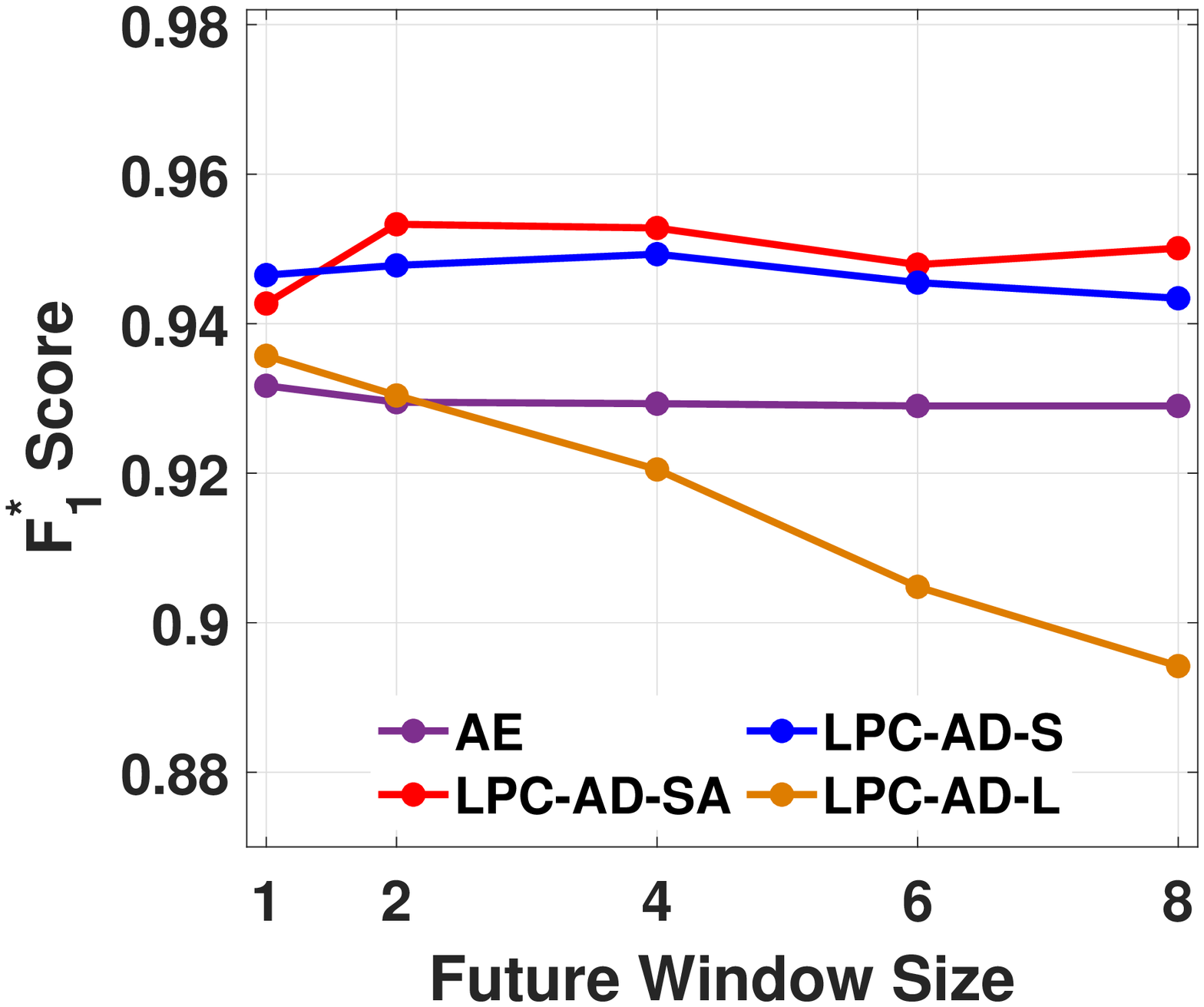}
		\label{fig:WindowSize2-F1}
	}
	\subfigure[AUROC score]{
		\centering
		\includegraphics[width=0.225\textwidth]{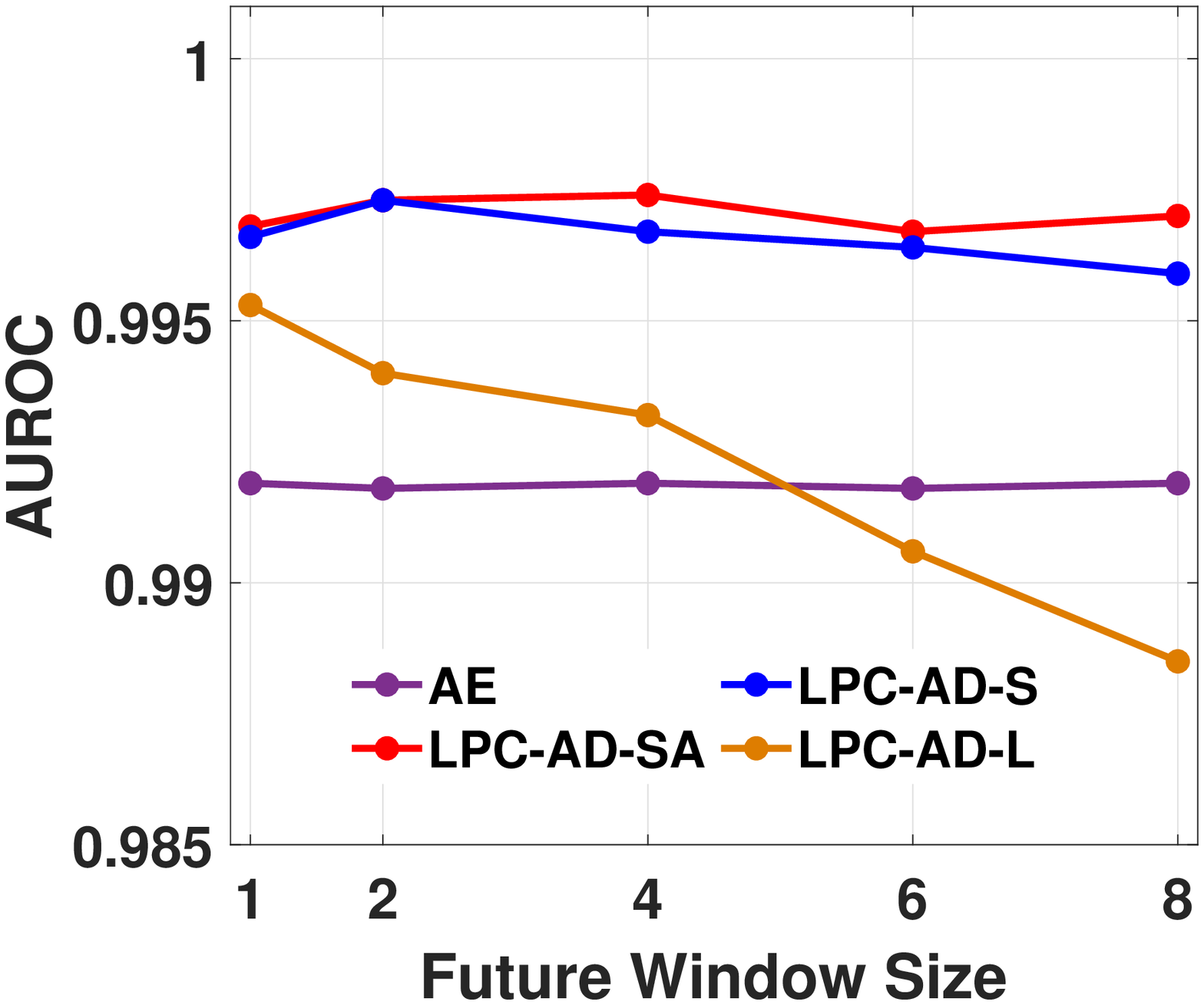}
		\label{fig:WindowSize2-AUROC}
	}
	\subfigure[Training time]{
		\centering
		\includegraphics[width=0.225\textwidth]{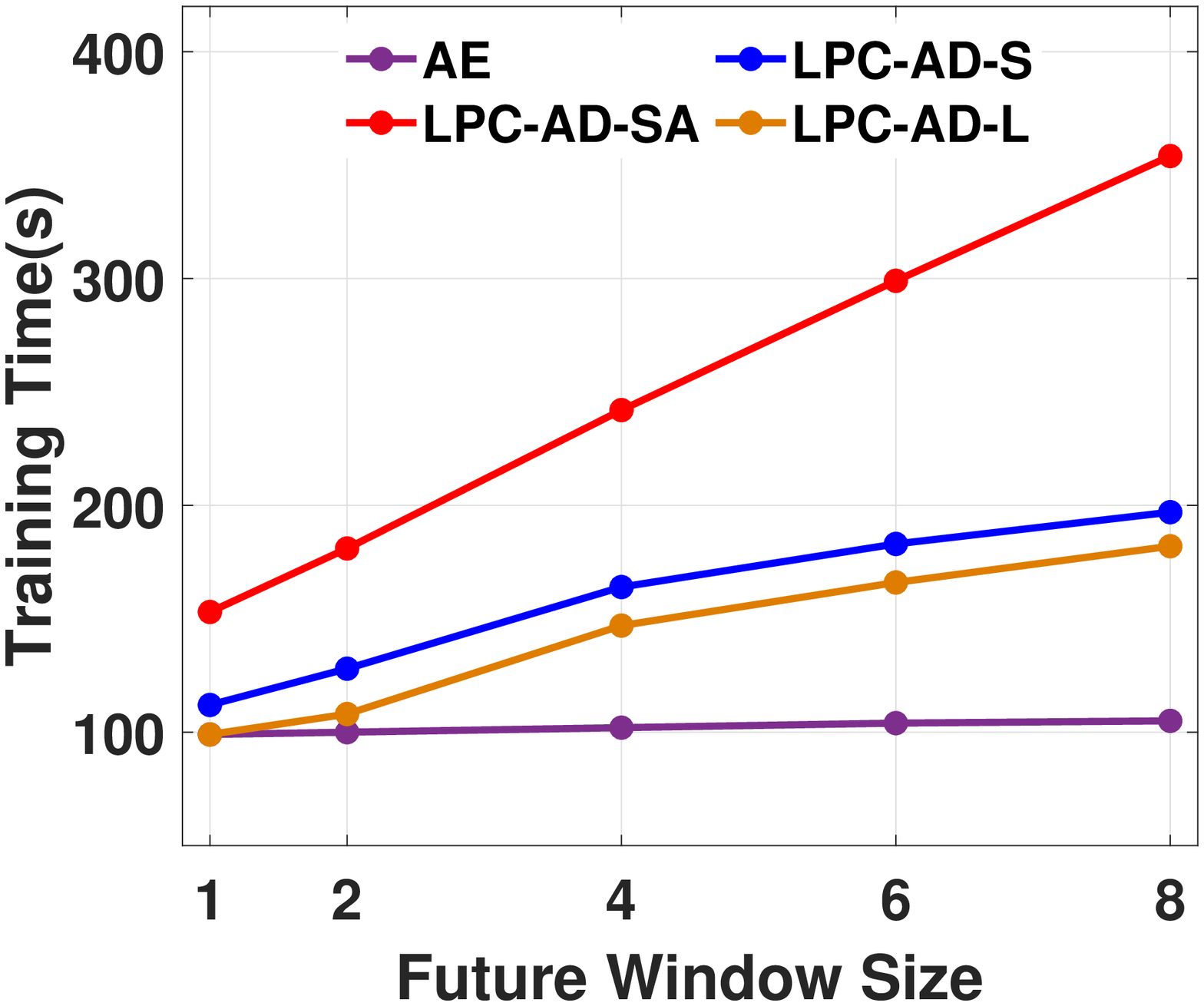}
		\label{fig:WindowSize2-TrainingTime}
	}
	\caption{Impact of future window size on the detection accuracy and training time.} 
	\label{fig:SMD Second window Size}
	\vspace{-0.12in}
\end{figure}

\noindent
{\bf $\bullet$ Impact of future window size.}  
Figure \ref{fig:SMD Second window Size} shows the impact of 
future window size $\ell$ on the detection accuracy and training time 
of three instances of LPC-AD and AE.  
From Figure \ref{fig:WindowSize2-OF1} one can observe that 
the $F_1$ curves of LPC-AD-SA, LPC-AD-S and AE are flat
as the historical window size $\ell_h$ 
increases from 1 to 8.  
Figure \ref{fig:WindowSize2-F1} and \ref{fig:WindowSize2-AUROC} show 
similar findings on the $F^\ast_1$ and AUROC metric. 
In summary, the detection accuracy of LPC-AD-SA, LPC-AD-S and AE 
is not sensitive to the historical window size.   
Figure \ref{fig:WindowSize2-OF1}-\ref{fig:WindowSize2-AUROC} 
show that the detection accuracy of LPC-AD-L decreases in 
future widow size.  
The reason that the linear predictor of LPC-AD-L under fits the data, 
making it less accurate for larger future window size.  
Figure \ref{fig:WindowSize2-TrainingTime} shows that the training time 
of LPC-AD-SA, LPC-AD-S and LPC-AD-L 
increases nearly linearly in the future window size $\ell$.  
Meanwhile, the training time of AE is almost unchanged, 
because the AE does not have the predictor component.   
These results show that three instances of LPC-AD scale well with respect 
to the future window size $\ell$.   
Lastly, Figure \ref{fig:SMD Second window Size} shows similar 
accuracy vs. training speed tradeoffs as Figure \ref{fig:SMD First window Size}, 
which is caused by capturing the nonlinear temporal dependence in the time series data.  

\begin{figure}[htb]
	\centering
	\subfigure[$F_1$ score]{
		\centering
		\includegraphics[width=0.225\textwidth]{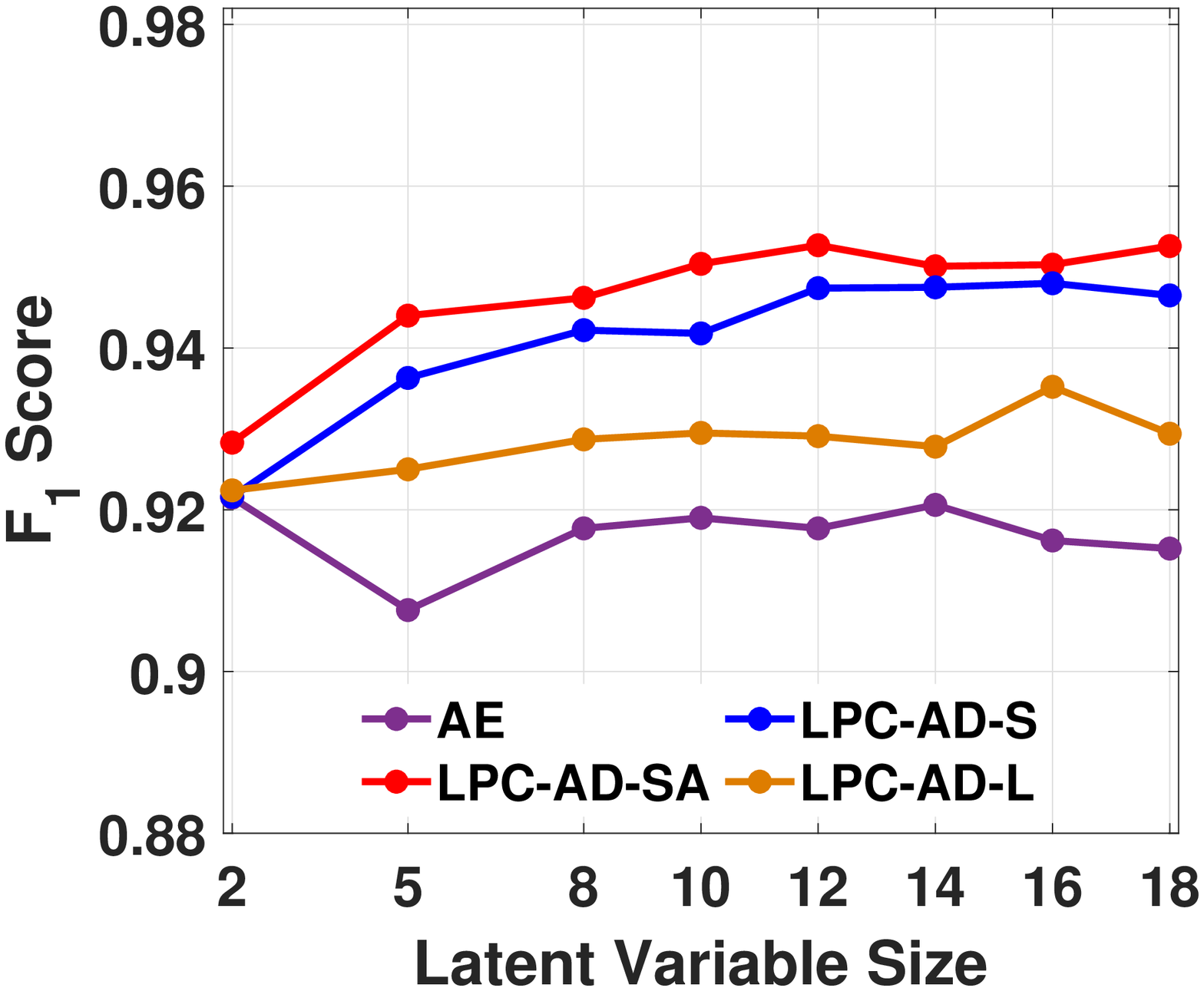}
		\label{fig:LatentVariableSize-OF1}
	}
	\subfigure[$F^*_1$ score]{
		\centering
		\includegraphics[width=0.225\textwidth]{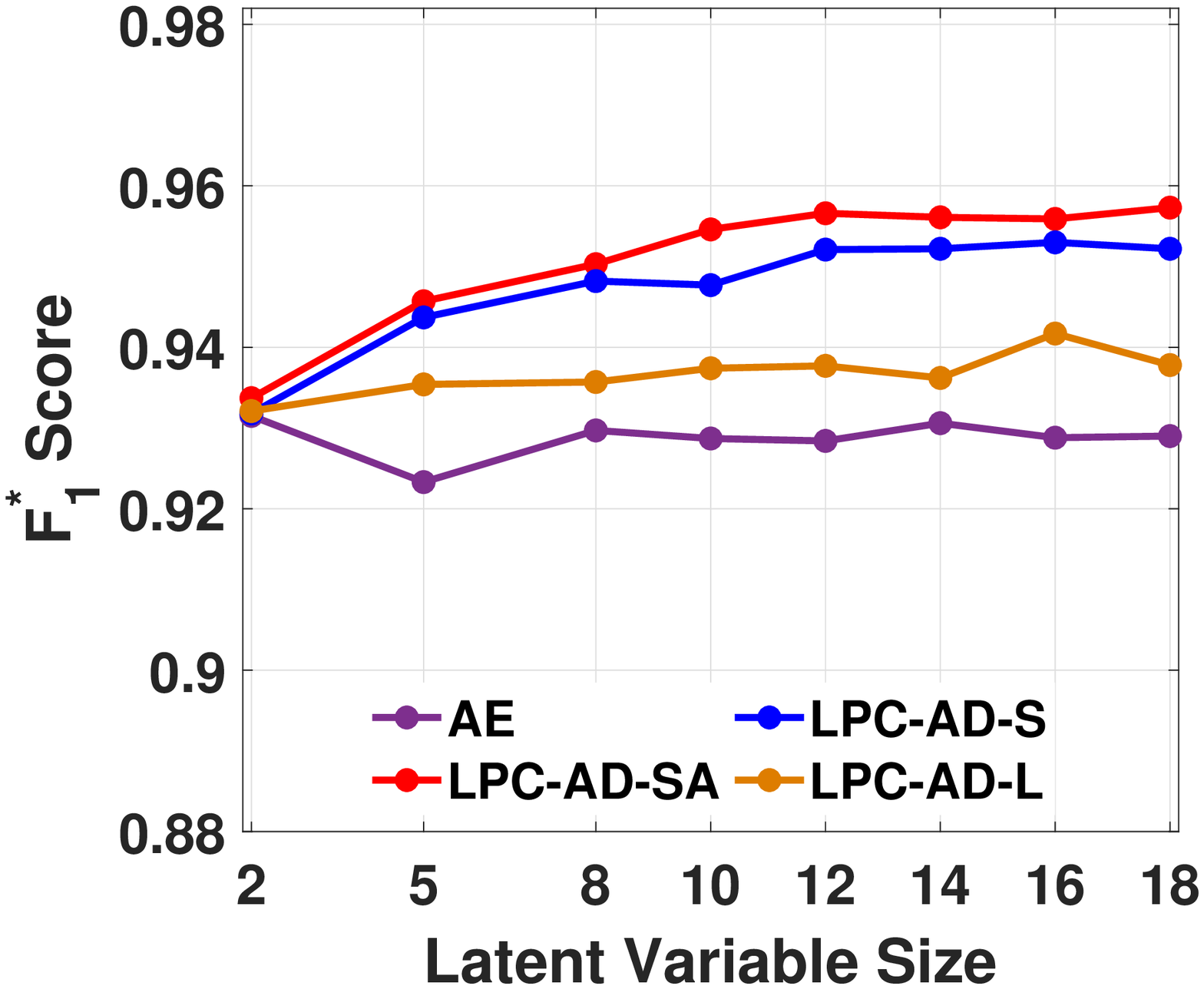}
		\label{fig:LatentVariableSize-F1}
	}
	\subfigure[AUROC score]{
		\centering
		\includegraphics[width=0.225\textwidth]{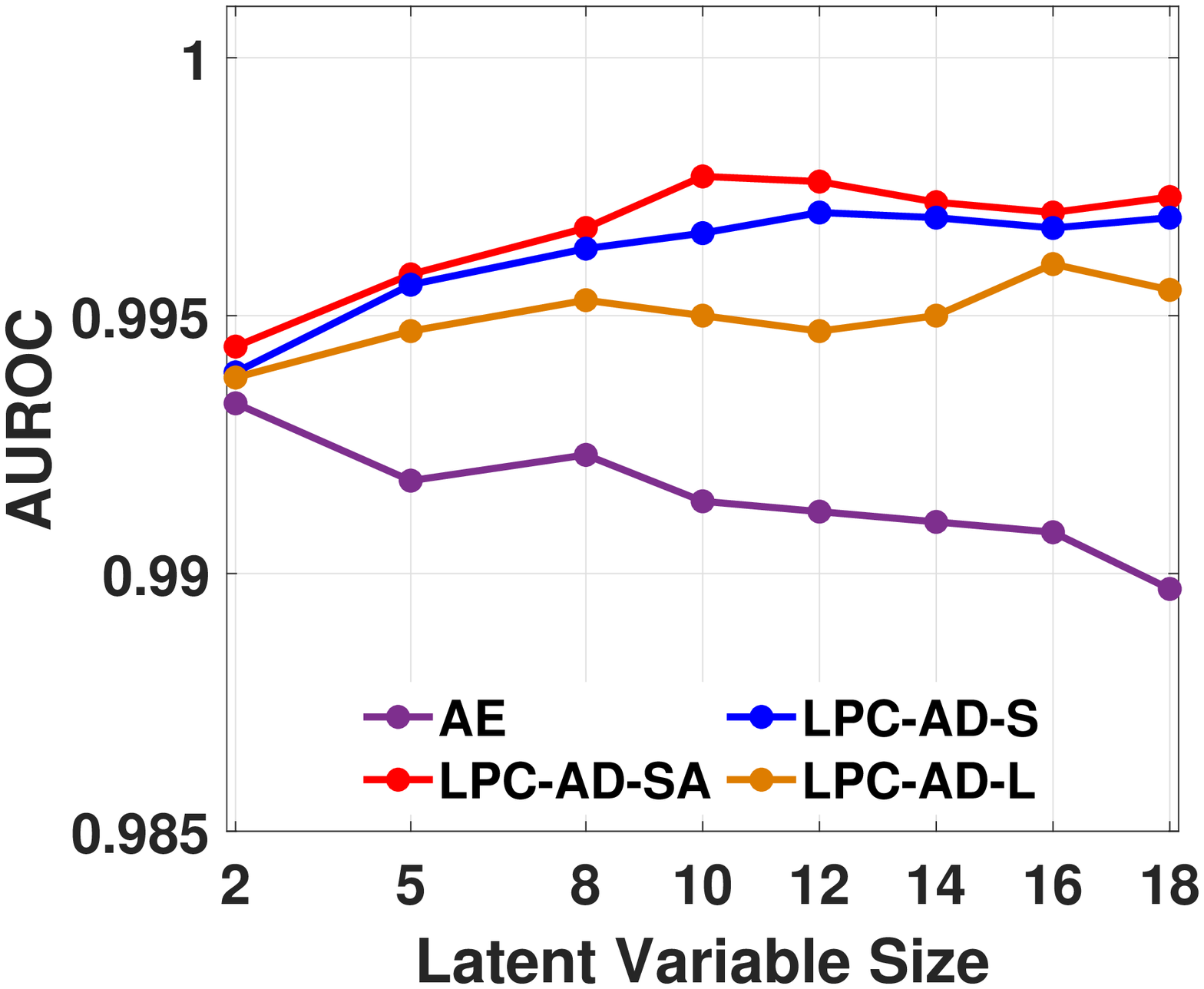}
		\label{fig:LatentVariableSize-AUROC}
	}
	\subfigure[Training time]{
		\centering
		\includegraphics[width=0.225\textwidth]{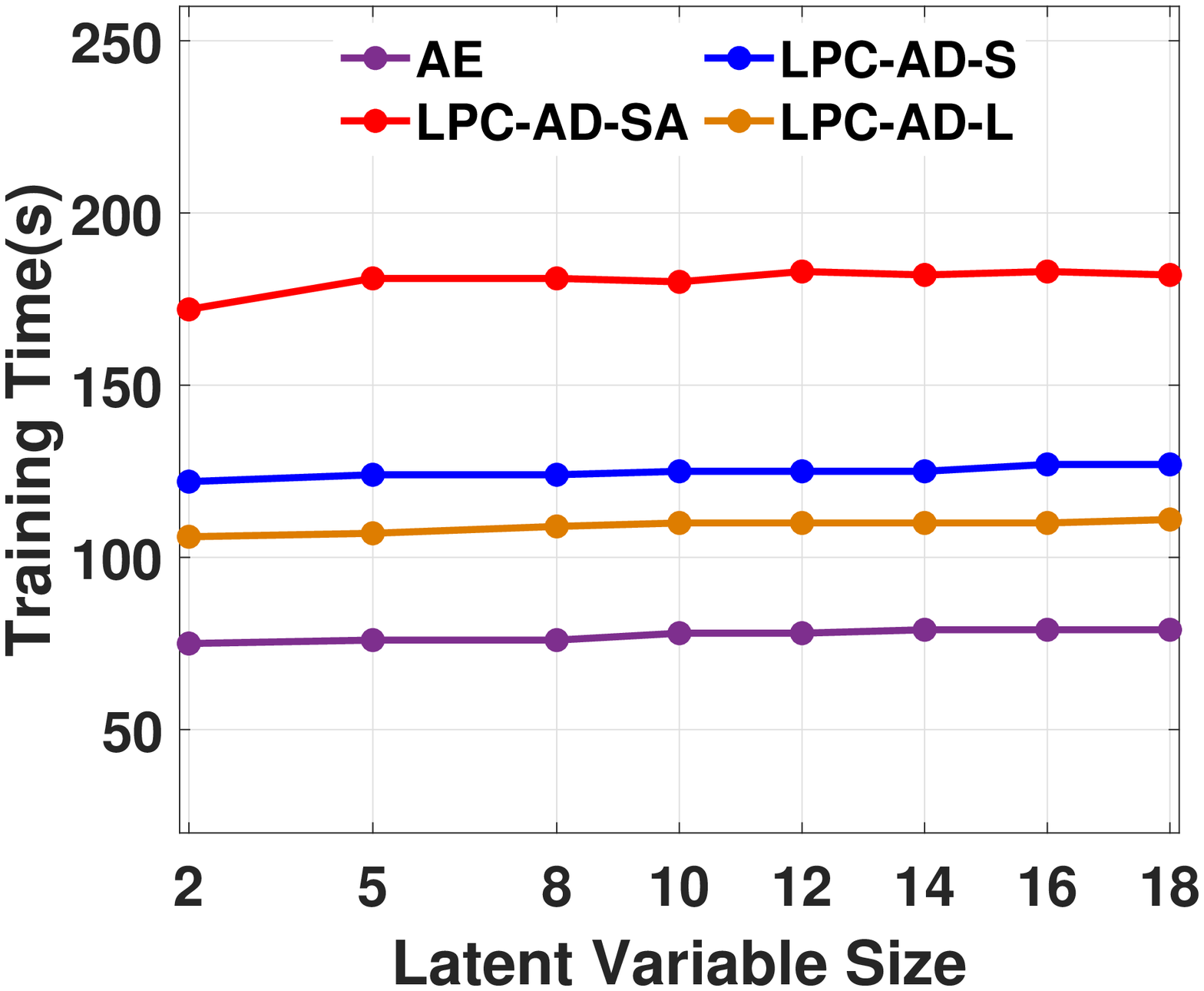}
		\label{fig:LatentVariableSize-TrainingTime}
	}
	\caption{Impact of latent variable dimension $N$ on the detection accuracy and training time.} 
	\label{fig:SMD Latent Variable Size}
\end{figure}

\noindent 
{\bf $\bullet$ Impact of latent variable dimension.}  
Figure \ref{fig:SMD Latent Variable Size} shows the impact of the dimension $N$ 
of latent variables on the detection accuracy and training time of the algorithms.
Figure \ref{fig:LatentVariableSize-OF1} shows that 
the $F_1$ scores of LPC-AD-SA and LPC-AD-S  
first increase slightly (less then 3\%) when the latent variable dimension increases 
from 2 to 12.  
Then they become flat when the latent variable dimension
 further increases from 12 to 18.  
The $F_1$ of LPC-AD-L and AE varies less than 1\% when 
the latent variable dimension increases from 2 to 18.  
Figure \ref{fig:LatentVariableSize-F1} shows similar findings 
on the $F^\ast_1$ metric.  
Furthermore, Figure \ref{fig:LatentVariableSize-AUROC} shows that 
the AUROC of LPC-AD-SA, LPC-AD-S, LPC-AD-L and AE varies less than 1\%
when the latent variable dimension increases from 2 to 18.  
In summary, the detection accuracy of LPC-AD-SA, LPC-AD-S and LPC-AD-L are 
not sensitive to the latent variable dimension.  
Figure \ref{fig:LatentVariableSize-TrainingTime} shows that the 
training time of LPC-AD-SA, LPC-AD-S, LPC-AD-L and AE 
are not sensitive to the dimesion of latent variables.  
Lastly, Figure \ref{fig:SMD Latent Variable Size} shows similar 
accuracy vs. training speed tradeoffs as Figure \ref{fig:SMD First window Size}, 
which is caused by capturing the nonlinear temporal dependence in MTS.  

\begin{figure}[h]
	\centering
	\vspace{-0.12in}
	\subfigure[$F^*_1$ score]{
		\centering
		\includegraphics[width=0.225\textwidth]{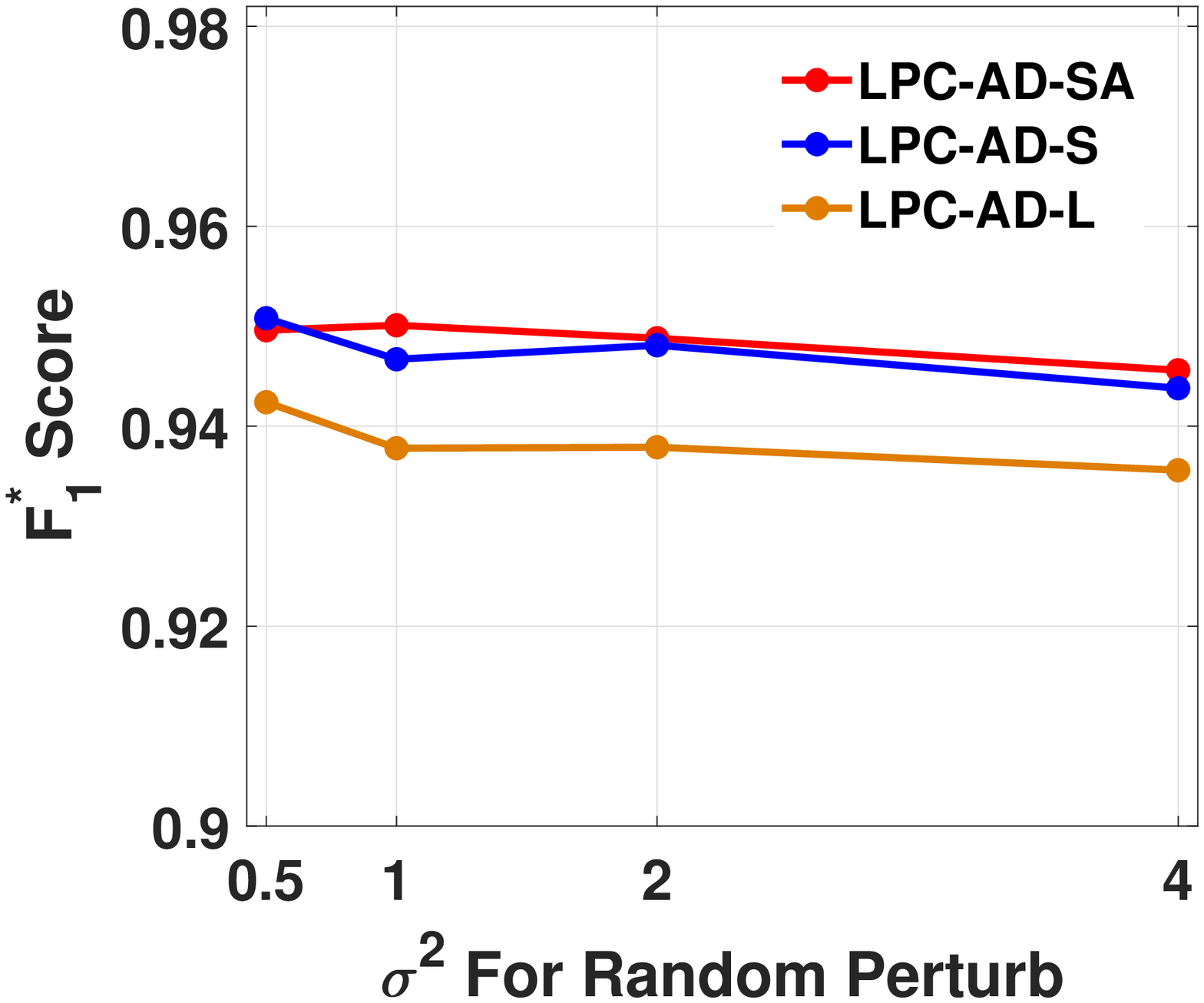}
		\label{fig:var-F1}
	}
	\subfigure[AUROC score]{
		\centering
		\includegraphics[width=0.225\textwidth]{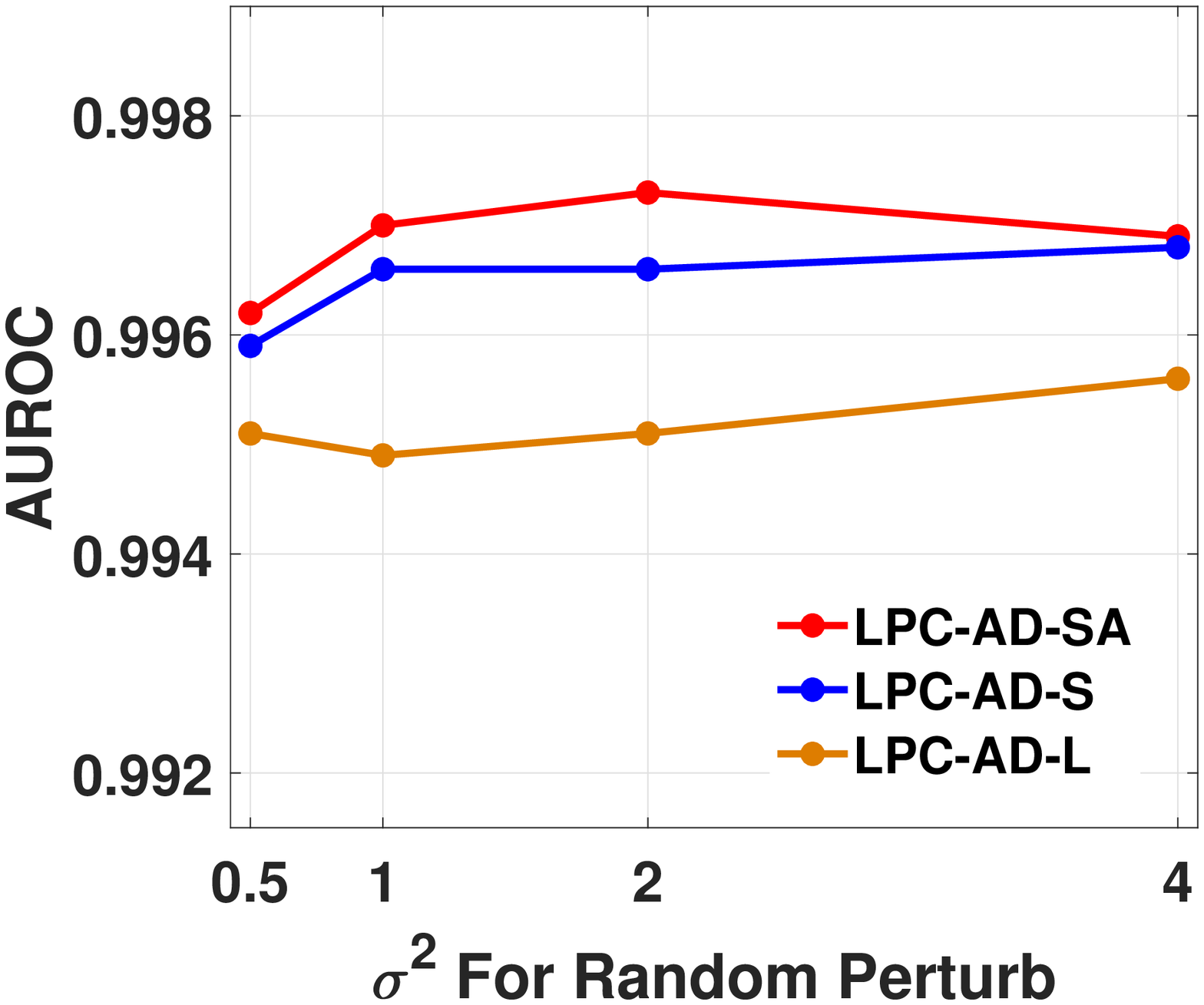}
		\label{fig:var-AUROC}
	}
	\caption{Impact of variance of random perturbation $\sigma^2$ on the detection accuracy and training time.} 
	\label{fig:SMD Variance} 
\end{figure}

\noindent 
{\bf $\bullet$ Impact of variance of random perturbation.}  
Note that $\bm{\Sigma}$ is the covariance matrix of the noise of random perturbation.  
For simplicity of presentation, we consider a special form of $\bm{\Sigma}$, i.e.,  
$
\bm{\Sigma} = \sigma^2 \bm{I}, 
$
where $\sigma^2$ measures the variance and $\bm{I}$ is 
an $N$ dimensional identity matrix.  
Figure \ref{fig:SMD Variance} shows the impact of variance $\sigma^2$ 
on the anomaly detection accuracy.  
We omit the experiment results on training time for brevity, 
as the variance has an ignorable influence the training time.  
Figure \ref{fig:SMD Variance} shows that the $F_1$ and AUROC of 
LPC-AD-SA, LPC-AD-S and LPC-AD-L vary slightly (less than 1\%) 
when the variance $\sigma^2$ increases from 0.5 to 4.  
This shows that LPC-AD-SA, LPC-AD-S and LPC-AD-L are not 
sensitive to the variance of the noise of random perturbation.   

\noindent{\bf Answer 3:}
\emph{
	Our LPC-AD-SA algorithm is robust to parameters of history window size, future window size, latent variable dimension and variance of random perturbation.
}

\subsection{Ablation Study}  

In this subsection, we conduct an ablation study to reveal a fundamental understanding 
on how two key components of our proposed algorithm, i.e., the predictor and 
the randomized perturbation operator, 
improve the anomaly detection accuracy.  

\noindent
{\bf Q4: Is it necessary to introduce the predictor and random perturbations?}  

\begin{figure}[h]
	\centering
	\subfigure[$F^*_1$ score]{
		\centering
		\includegraphics[width=0.225\textwidth]{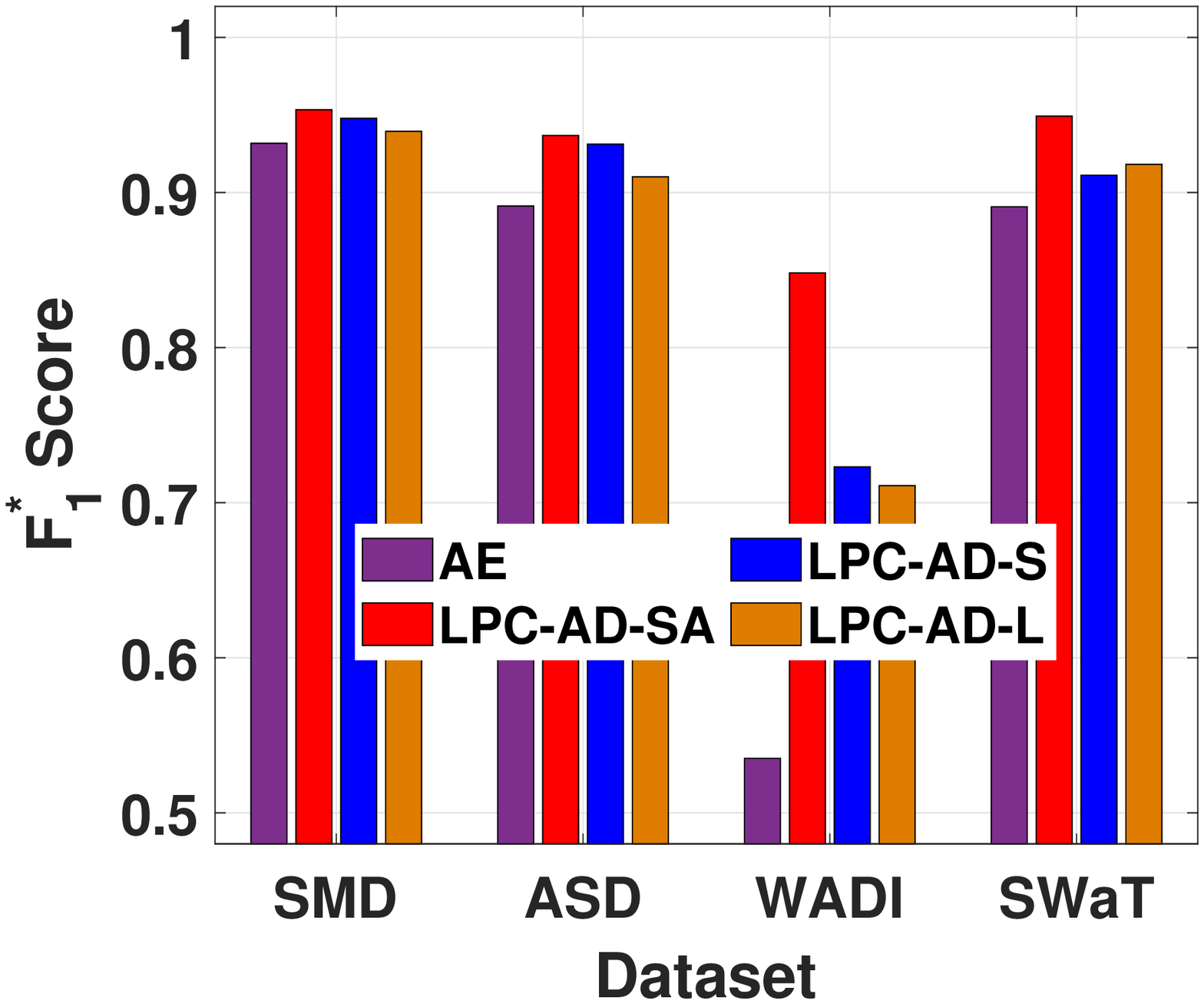}
		\label{fig:Images/AblationStudy/Ablation-F1}
	}
	\subfigure[AUROC score]{
		\centering
		\includegraphics[width=0.225\textwidth]{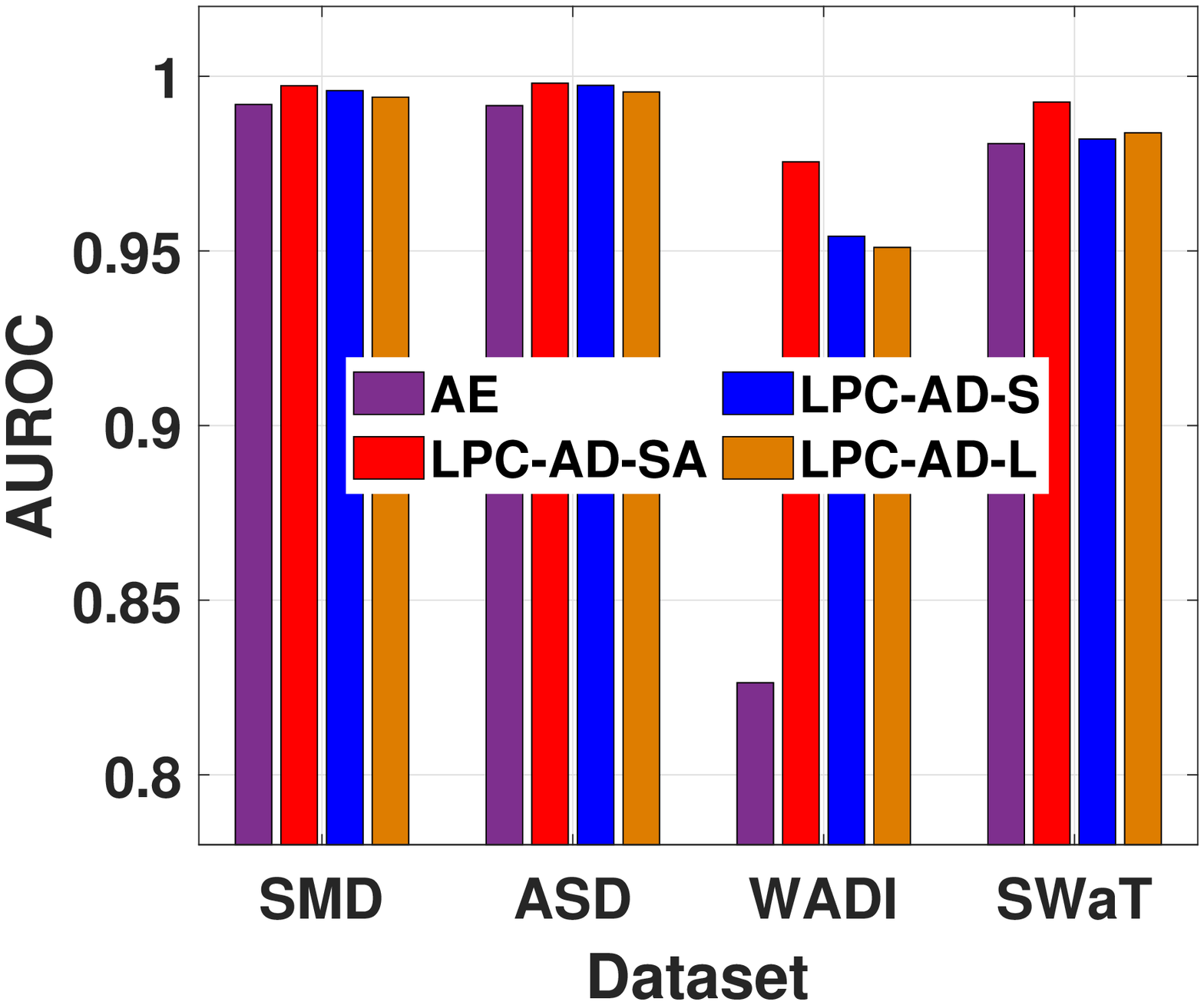}
		\label{fig:Images/AblationStudy/Ablation-AUROC}
	}
	\caption{Impact of the predictor on the anomaly detection accuracy.} 
	\label{fig:Ablation Study}
	\vspace{-0.12in}
\end{figure}

\noindent
{\bf $\bullet$ Impact of the predictor.}  
Figure \ref{fig:Ablation Study} shows the $F_1$ and AUROC of 
AE and instances of LPC-AD with different predictors.  
Note that the AE corresponds to a degenerated variant of LPC-AD 
by deleting the predictor component.  
In addition, AE does not have the randomized perturbation operator, 
as it is built on the predictor.  
Figure \ref{fig:Ablation Study} shows that the 
$F_1$ and AUROC of three instances of LPC-AD are always higher than AE 
over four datasets.  
Particularly on the WADI dataset, the LPC-AD-L improves the $F_1$ and AUROC of AE drastically.  
This shows that even a simple linear predictor can improve the detection accuracy significantly.  
One can also observe that the LPC-AD-S has a higher $F_1$ and AUROC than LPC-AD-L 
in most cases.  
This shows the importance of capturing nonlinearity in the prediction.   
Moreover, the LPC-AD-SA has a higher $F_1$ and AUROC than LPC-AD-S 
on all four datasets.  
In summary, there is rich temporal dependence among the latent variables  
and it is important to use non-linear predictor to capture it well in order to improve
 the anomaly detection accuracy.  

\noindent
{\bf $\bullet$ Impact of randomized perturbation.}  
To study the impact of randomized perturbation, 
we consider a variant of LPC-AD without randomized perturbation denoted by LPC-AD-N,  
where N refers to no randomized perturbation.  
In particular, LPC-AD-N is obtained by setting the randomized perturbation operator 
in the following deterministic form
$
\texttt{RandPerturb}({\bm Z}_{t+1}, \widehat{{\bm Z}}_{t+1}, \bm{\epsilon}; \bm{\Theta}_{\text{RP}}) 
= \widehat{{\bm Z}}_{t+1}.  
$
Figure \ref{fig:Images/AblationStudy/Ablation-F1} shows that LPC-AD-SA always have a higher 
$F_1$ than LPC-AD-N, where the improvement reaches a significant number of 5\%.  
Figure \ref{fig:Images/AblationStudy/Ablation-AUROC} shows that LPC-AD-SA always has a higher 
AUROC than LPC-AD-N, except on the WADI dataset.  
On the WADI dataset, the AUROC of LPC-AD-SA is around 0.5\% lower than LPC-AD-N.

\begin{figure}[h]
	\centering
	\subfigure[$F^*_1$ score]{
		\centering
		\includegraphics[width=0.225\textwidth]{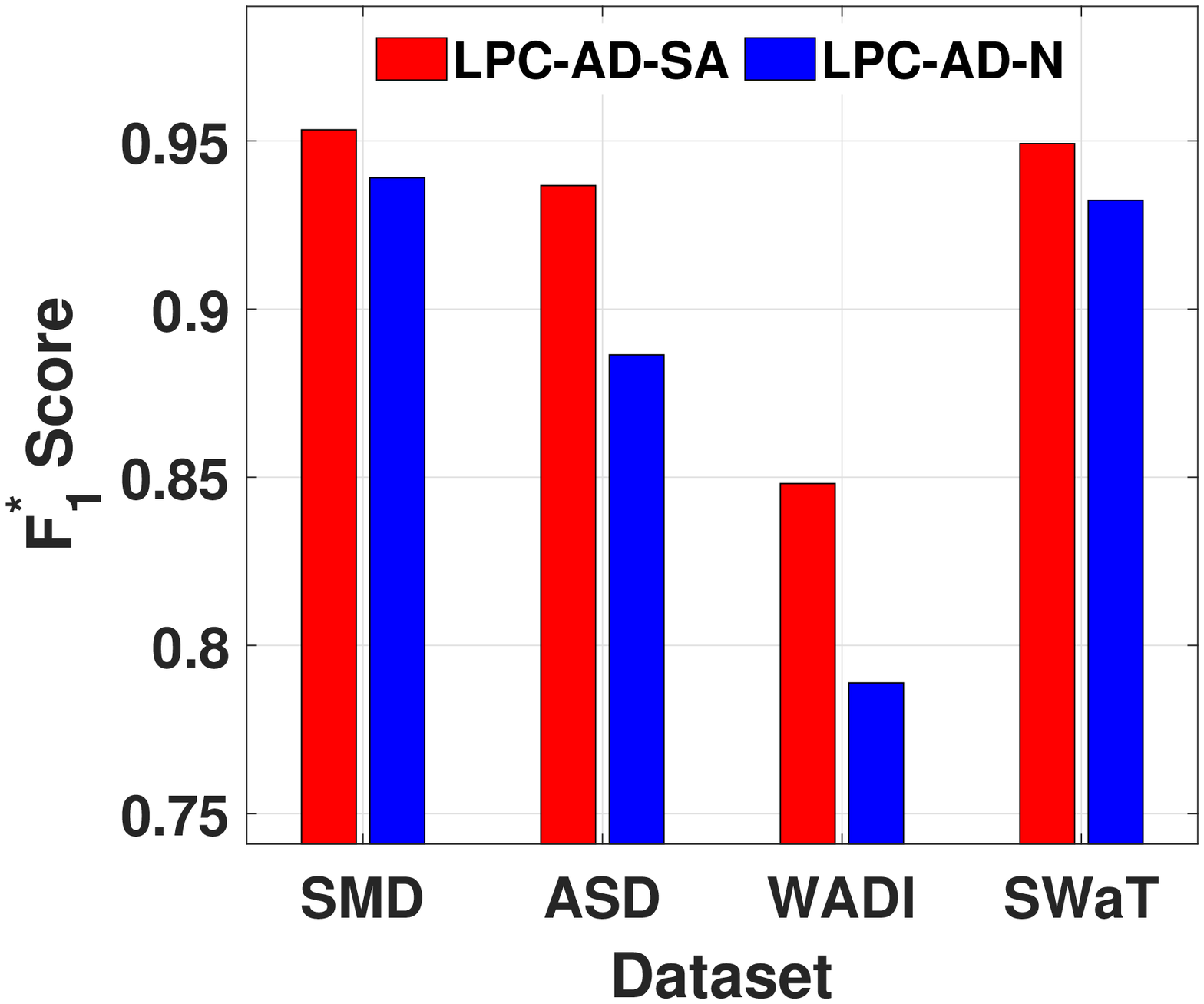}
		\label{fig:Images/AblationStudy/Ablation-F1}
	}
	\subfigure[AUROC score]{
		\centering
		\includegraphics[width=0.225\textwidth]{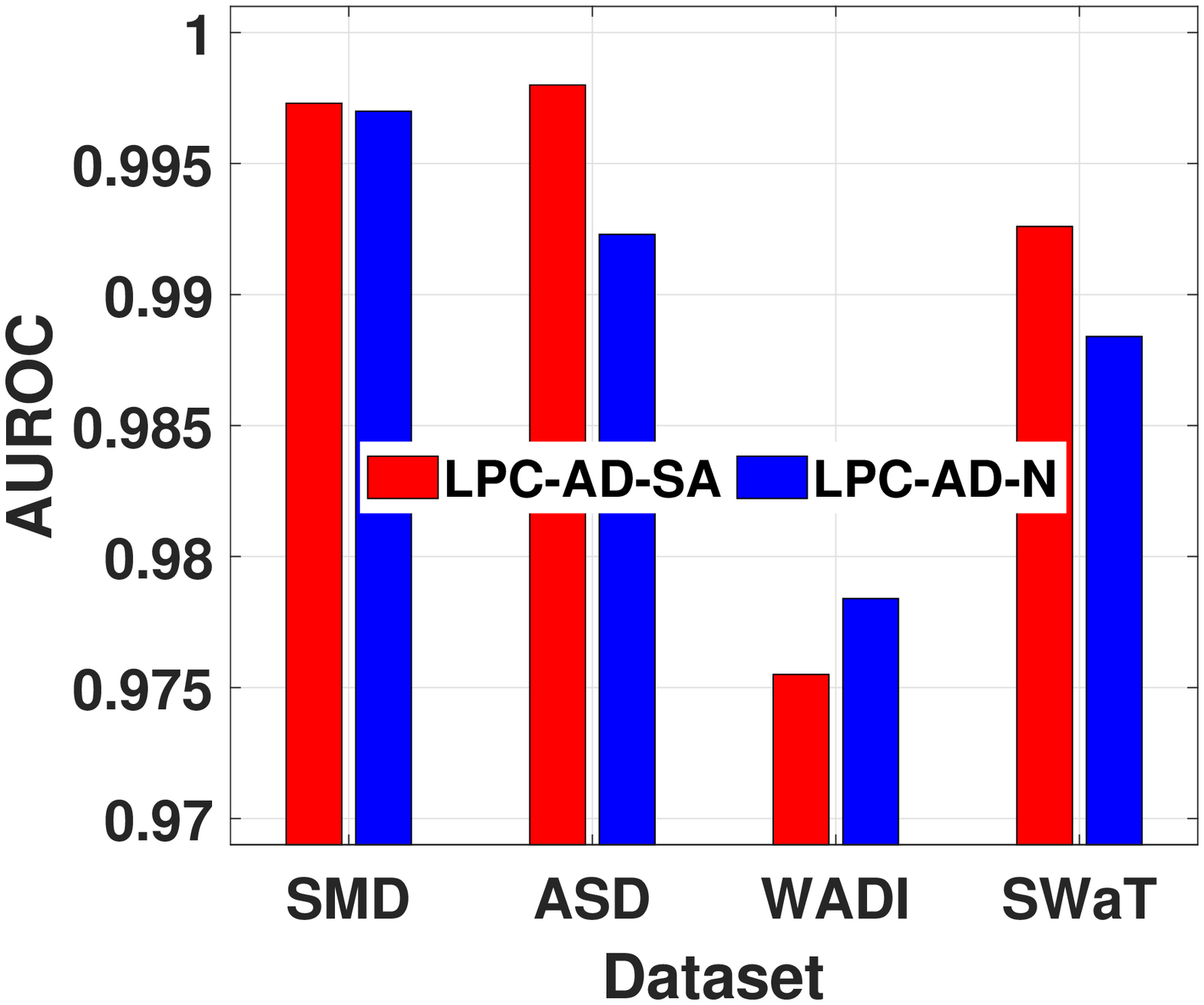}
		\label{fig:Images/AblationStudy/Ablation-AUROC}
	}
	\caption{Impact of randomized perturbation on the detection accuracy.} 
	\label{fig:Ablation-Random Study}
	\vspace{-0.12in}
\end{figure}

\noindent{\bf Answer 4:}
\emph{
	Adding the predictor can increase the F1 score by as high as 9.16\%. 
	Adding the randomized perturbation can increase the F1 score by as high as 5\%.
}
\section{Conclusion}
This paper presents \texttt{LPC-AD}, 
which has shorter training time 
than SOTA deep learning methods 
that focus on reducing training time and it 
has a higher detection accuracy than 
SOTA sophisticated deep learning methods that 
focus on enhancing detection accuracy.  
These merits of \texttt{LPC-AD} are supported by novelty in the design 
and extensive empirical evaluations.  
In particular, \texttt{LPC-AD} 
contributes a generic architecture \texttt{LPC-Reconstruct}, 
which is a novel combination of ideas 
from autoencoder, 
predictive coding  
and randomized perturbation.    
We also contribute a new randomized perturbation method to avoid 
overfitting of anomalous dependence patterns.  
Extensive experiments validate superior performance and 
reveal fundamental understating of our method. 

\bibliographystyle{abbrv}
\bibliography{reference}

\end{document}